\documentclass[11pt]{article}
\usepackage[]{ACL2023}

\usepackage{fancyhdr}
\fancypagestyle{plain}{%
  \fancyhf{}%
  \fancyhead[C]{Accepted to EACL 2026 Demo Track (Camera-Ready Version)}%
  \fancyfoot[C]{\thepage}%
}
\fancypagestyle{fancydefault}{%
  \fancyhf{}%
  \fancyhead[C]{Accepted to EACL 2026 Demo Track (Camera-Ready Version)}%
  \fancyfoot[C]{\thepage}%
}
\usepackage{natbib}
\usepackage{times}
\usepackage{latexsym}
\usepackage{multirow}
\usepackage{booktabs}
\usepackage{tabularx}
\usepackage{amssymb}
\usepackage{comment}
\usepackage{gensymb}
\usepackage{microtype}
\usepackage{comment}
\usepackage{color}
\usepackage{amsmath} 
\usepackage{amsfonts} 
\usepackage{inconsolata}
\usepackage{graphicx}
\usepackage{rotating}
\usepackage{bbding}
\usepackage{pifont}
\usepackage{wasysym}
\usepackage{amssymb}
\usepackage{listings}
\usepackage{array} 
\usepackage{subfig}
\usepackage{enumitem}
\usepackage{booktabs}   
\usepackage{multirow}   
\usepackage{makecell}   
\usepackage[normalem]{ulem}
\usepackage{changepage}
\usepackage[T1]{fontenc}

\usepackage[utf8]{inputenc}
\usepackage[ruled,vlined]{algorithm2e}
\usepackage{microtype}
\usepackage{placeins} 
\usepackage{changepage}
\usepackage{fancyvrb}
\usepackage{inconsolata}
\usepackage[most]{tcolorbox}
\tcbuselibrary{listings,breakable}

\newtcblisting{myverbatim}{
      arc=3mm,
      top=1mm,
      bottom=1mm,
      left=1mm,
      right=1mm,
      boxrule=0.5pt,
      colback=gray!5,
      colframe=black,
      listing only,
      breakable,
}
\usepackage{graphicx}
\newcommand{\cmark}{\ding{51}}%
\newcommand{\xmark}{\ding{55}}%
\usepackage{enumitem}

\definecolor{customblue}{RGB}{33,167,234}

\title{Using a Human-AI Teaming Approach to Create and Curate Scientific Datasets with the \textsc{SciLire} System}

\author{Necva B{\"o}l{\"u}c{\"u}$^{1\dagger}$, Jessica Irons$^{1\dagger}$, 
Changhyun Lee$^{1}$, Brian Jin$^{1}$, \\ 
\textbf{ Maciej Rybinski$^{2}$\thanks{This work was done when the author was affiliated with CSIRO Data61.}, Huichen Yang$^{1}$, Andreas Duenser$^{1\dagger}$, Stephen Wan$^{1}$\thanks{Primary authors for this work.}} \\
  $^{1}$CSIRO, Sydney, Australia \\
  \texttt{firstname.lastname@csiro.au} \\
  $^{2}$ITIS, University of Málaga, Málaga, Spain\\
  \texttt{maciek.rybinski@uma.es}}

\begin{document}
\maketitle

\thispagestyle{plain} 
\begin{abstract}

The rapid growth of scientific literature has made manual extraction of structured knowledge increasingly impractical. To address this challenge, we introduce \textsc{SciLire}, 
a system for creating datasets from scientific literature.  \textsc{SciLire} has been designed around Human-AI teaming principles centred on workflows for verifying and curating data.
It facilitates an iterative workflow in which researchers can review and correct AI outputs.  Furthermore, this interaction is used as a feedback signal to improve future LLM-based inference.
We evaluate our design using a combination of intrinsic benchmarking outcomes together with real-world case studies across multiple domains.
The results demonstrate that \textsc{SciLire} improves extraction fidelity and facilitates efficient dataset creation.

\end{abstract}

\section{Introduction}


The exponential growth of scientific literature, which makes it increasingly challenging for researchers to stay up to date with the latest scientific developments~\citep{cai2024uni, reddy2025towards}, represents an opportunity: scientific papers can be mined to generate high-value datasets~\citep{dunn2022structured, jiang2025enzyme, wei2025finding}. Such datasets are key in creating Artificial Intelligence (AI) to revolutionise scientific workflows and discovery, an endeavour generally referred to as \textit{AI for Science} (AI4S).  

Building on this potential, recent progress in Large Language Models (LLMs) offers powerful new tools, such as Elicit\footnote{\url{https://elicit.com/}} and SciSpace\footnote{\url{https://scispace.com}}, to assist researchers in navigating and extracting knowledge from the vast scientific literature. 
However, these systems treat AI-data extraction as a single pass.  Given that AI results are usually not perfect, single-pass tools force users to improve data outside of the tool without AI-assistance, a challenge when working with big datasets.
This can limit the adoption of such tools in research workflows where output must conform to a certain standard and where such user processes to validate data manually are often arduous and time-consuming~\citep{10.1145/3491102.3502068,pham2025collaboration, schmidt2025data}.


Indeed, recent studies highlight risks related to LLM-based extraction: these models may generate hallucinated (confabulated) information, with empirical evaluations showing nontrivial error rates that require human correction and verification~\citep{helms2025using}. Reviews of AI for literature synthesis further highlight ongoing problems with explainability and reliability, showing that generative AI cannot be fully trusted without expert oversight~\citep{bolanos2024artificial}. 

We adopt a Human-AI Teaming (HAT)~\cite{10.3389/frai.2023.1250725} design in \textsc{SciLire}, enabling users to curate data (hereafter: \textbf{HAT} for \textbf{D}ata \textbf{C}uration (HAT-DC)).
By combining expert validation with AI-assisted extraction, researchers can correct errors and mitigate hallucinations.
Moreover, iterative human feedback helps improve model performance and fosters transparency, accountability, and trust in AI-enabled workflows~\citep{gao2025agent, schroeder2025largelanguagemodelshumanintheloop}.

\begin{table}[t]
\resizebox{\columnwidth}{!}
{
\begin{tabular}{r cccc}\toprule
\bf Tool 
& \bf \begin{tabular}[c]{@{}c@{}}Dynamic\\Sampling\end{tabular} 
& \bf \begin{tabular}[c]{@{}c@{}}Multi-record\\Support\end{tabular} 
& \bf \begin{tabular}[c]{@{}c@{}}Provenance\\Data\end{tabular} 
& \bf \begin{tabular}[c]{@{}c@{}}Table Export\\(CSV,  JSON, $\cdots$)\end{tabular} 
\\
\midrule
Elicit 
& \xmark\textsuperscript{$\ddagger$}  
& \xmark 
& \cmark\textsuperscript{$\dagger$}  
& \cmark 
\\
SciSpace 
& \xmark\textsuperscript{$\ddagger$}  
& \xmark 
& \cmark\textsuperscript{$\dagger$}   
& \cmark 
\\
NotebookLM 
& \xmark\phantom{$\dagger$} 
& \cmark  
& \cmark\textsuperscript{$\dagger$} 
& \xmark 
\\
Claude.ai 
&  \xmark\phantom{$\dagger$} 
& \cmark 
& \xmark\phantom{$\dagger$}  
&  \cmark 
\\
\midrule
\textsc{SciLire}(Ours) & \cmark\phantom{$\dagger$} 
& \cmark 
& \cmark\textsuperscript{$\dagger$} 
& \cmark 
\\
\bottomrule
\end{tabular}
}
\caption{Comparison of data curation tools' functionality, highlighting differences from \textsc{SciLire}. \textsuperscript{$\ddagger$} Elicit and SciSpace can be made to accept static examples in column definitions. \textsuperscript{$\dagger$} Elicit, SciSpace and NotebookLM provide paragraph- or sentence-level citations. \textsc{SciLire} provides the degree of alignment with the source, along with relevant paragraphs. 
\textbf{Multi-record support} indicates if a tool is designed to produce multiple extracted records per document. 
}
\label{tab:related_work}
\end{table}

\textsc{SciLire} differs from existing tools in that it supports iterative extraction and correction, enables dynamic sampling using the user's curation history as an evolving source of examples, and scales to large literature search collections.

We evaluate the HAT-DC design of \textsc{SciLire} using public scientific datasets, provide real-world case studies, and report on feedback from users. Our contributions are: (1) evaluation of an HAT-DC workflow; (2) insights into the effectiveness of dynamic sampling.



\section{Related Work}
\label{sec:related}



Our work focuses on AI-powered tools leveraging LLMs for data curation, such as Elicit and SciSpace, which extract key information from PDFs. 
A listing of existing tools that extract some data from PDFs is presented in  Table~\ref{tab:related_work}. 
For the HAT-DC, two key features are required: (1) the exporting of tables (in CSV or JSON), and (2) provision of provenance data for data verification and curation -- this leaves Elicit and SciSpace as the two most relevant tools to our workflow.

Beyond these tools, there is a growing body of research that looks more broadly at how information from scientific papers can be turned into structured tables. ArxivDIGESTables~\citep{newman-etal-2024-arxivdigestables} studies cross-paper table generation with LLMs and proposes an automatic evaluation method, while ArXiv2Table~\citep{deng-etal-2024-text} presents a more comprehensive benchmark in the computer science domain. Several domain-specific efforts have also attempted to extract structured or tabular information directly from research papers in other domains, such as material science, chemistry and food manufacturing~\citep {dunn2022structured, wei2025finding, bolucu2025evaluation}. Collectively, these studies highlight growing interest in turning unstructured documents into tabular formats, a goal that AI-powered tools put into practice.


\section{AI-augmented Curation Workflow}

\textsc{SciLire} is designed to support the existing data curation workflow, with the use of AI focused on human skill augmentation. Typically, users iterate through possible schema structures (either provided by the user or selected from built-in templates) with \textsc{SciLire}.
%
The HAT-DC workflow is as follows:\footnote{See Appendix~\ref{sec:appendix_demo_walkthrough} for screenshots and a more detailed system usage description.}
\begin{enumerate}[noitemsep]
\item \textbf{Bibliography Upload and Schema Definition:} Users upload a collection of documents and provide a schema file (e.g., a spreadsheet) specifying the column headers of the target curated table. The schema can be modified at any time. Once the documents are uploaded, \textsc{SciLire} automatically triggers the document preprocessing module to provide machine-readable versions of the text.

\item \textbf{Pilot Phase:} Users select a small sample of documents ($\le$10), generate data tables, and manually vet and correct the outputs. During this process, they also assess which documents and extracted records are relevant to the target aim.  For relevant documents, users verify the generated results, correcting and curating data where necessary. Verified corrections are used by the system for dynamic sampling for subsequent batches. The pilot phase (multiple batches) is repeated for $\sim$ 50–100 documents or until the user is satisfied.
\item \textbf{Batch Phase:} The remaining documents are processed at scale.  The user either continues to check data or exports the data as is.\footnote{The degree of further verification will depend on thresholds for acceptable quality related to the user's end goal.}
\end{enumerate}

\begin{figure*}
    \centering
    \includegraphics[width=\linewidth]{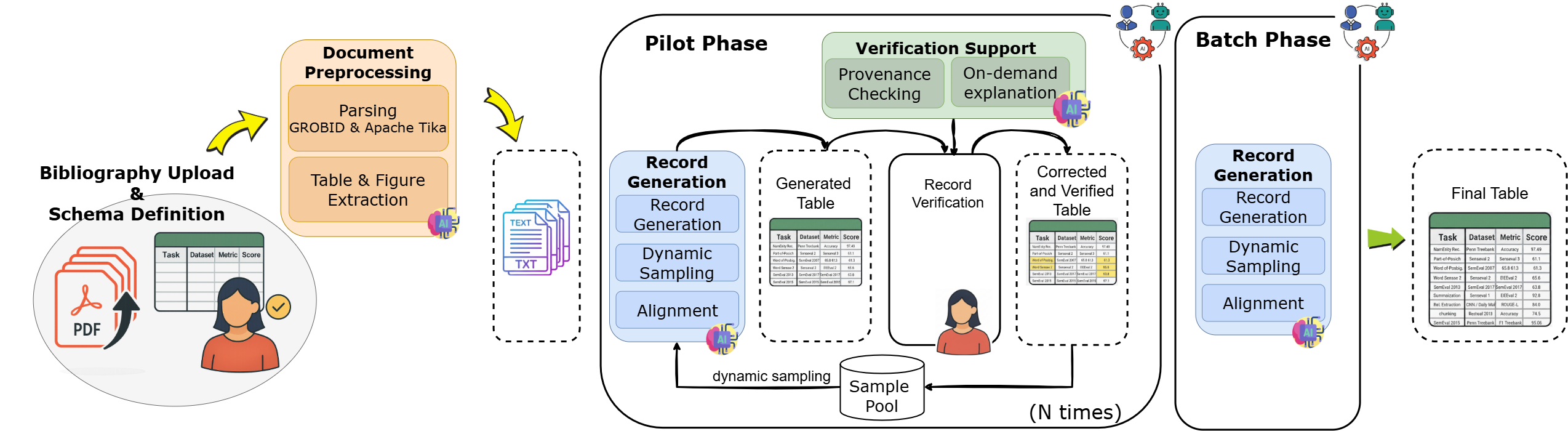}
    \caption{\textsc{SciLire} components and AI-augmented curation workflow.}
    \label{fig:scilire-architecture}
\end{figure*}

\section{System Components}\label{sec:system_components}
\textsc{SciLire} is built as a modular framework for data curation. Its architecture (Figure~\ref{fig:scilire-architecture}) comprises three modules that handle document pre-processing, LLM-based record generation, and verification support. Together, these modules form a flexible and transparent pipeline designed to support the user in a HAT-DC workflow.

\subsection{Document Preprocessing}\label{ssec:document_processing}
\paragraph{Parsing.}
\textsc{SciLire} checks each document for PDF type. If the PDF contains machine-readable text, it is processed using two PDF parsing pipelines (\texttt{GROBID}~\citep{lopez2009grobid}\footnote{\url{https://github.com/kermitt2/grobid}} and \texttt{Apache Tika}\footnote{\url{https://tika.apache.org}}).  For PDFs with no machine-readable text (e.g., PDFs with only scanned images), \textsc{SciLire} uses the \texttt{OCRmyPDF}\footnote{\url{https://github.com/ocrmypdf/OCRmyPDF}} library to convert a PDF of page images into a PDF with machine-readable text. The resulting PDFs are processed using the same pipelines as non-scanned PDFs, ensuring consistency across all documents: (1) The \texttt{GROBID} is the preferred pipeline, providing high-quality text extraction and structured metadata~\citep{meuschke2023benchmark}, particularly for scientific literature. The \texttt{Tika} pipeline provides an alternative when \texttt{GROBID} occasionally fails, thereby providing better support of other text genres. Since \texttt{GROBID} does not process figures and provides limited table extraction quality~\citep{meuschke2023benchmark}, we develop a custom table and figure extraction module.

\paragraph{Table \& figure extraction.}\label{paragraph:table_extraction}
We implemented a two-stage pipeline (Figure~\ref{fig:tr_pipeline}, Appendix~\ref{sec:appendix_table_recognition}) for the task. First, \textsc{SciLire} detects pixel regions indicating tables and figures in PDFs.\footnote{Currently, we do not further process figures.} \textsc{SciLire} then performs table structure and caption recognition on the detected regions. Finally, the inferred table structure, together with table contents and captions, is rendered in markdown format. This is appended to the text extracted by GROBID based on the position of tables in the PDF. 
Implementation details of this module are provided in the Appendix~\ref{sec:appendix_table_recognition}.

\paragraph{Chunking.} Since LLMs have fixed context windows, long documents are segmented into overlapping chunks. We apply a configurable sliding-window strategy (by characters), preserving local coherence through overlapping spans. The window size and overlapping ratio are configurable parameters in the system (window size=LLM context length, overlap=10\%).

\subsection{Record Generation Module}\label{ssec:record_generation}
Given a user's schema that outlines concepts of interest (e.g., context or variables in an experiment with a measured result), \textsc{SciLire} automatically constructs prompts to generate structured records using an LLM. Initially, the prompt follows a zero-shot baseline approach. If human-corrected data exists, the prompt uses a few-shot ``In-Context Learning'' (ICL) approach~\citep{ghosh-etal-2024-toward}, which dynamically picks an example to include in the prompt.

\textsc{SciLire} uses the schema concepts to define a JSON dictionary structure~\citep{oestreich2025evaluatingstructureddecodingtexttotable} as the desired output format, which is included in the prompt (Appendix~\ref{sec:appendix_prompts}).  This structure also houses any ICL examples if required.  
Two versions of the prompt are then created, using data from the GROBID and Tika pipelines.  The two prompts are then sent to the LLM.\footnote{\textsc{SciLire} can use any LLM, including closed and open weight models. In the \textsc{SciLire}, we use \texttt{GPT-4o} as the model.} 
%

%

\paragraph{Alignment.} 
Since the generation phase can yield two record sets (GROBID-based and Tika-based) per PDF, we merge the sets using the Hungarian maximum-matching algorithm~\citep{kuhn1955hungarian} to identify overlapping records. We compute a similarity matrix by encoding records with sentence embeddings\footnote{\url{https://huggingface.co/sentence-transformers/all-MiniLM-L6-v2}} and computing pairwise cosine similarity.
The algorithm selects the optimal one-to-one alignment, allowing \textsc{SciLire} to suggest alternative records which users can compare during the curation task.

\paragraph{Dynamic sampling for ICL.}
Selecting an effective ICL example is critical in few-shot prompting. Instead of using static examples, \textsc{SciLire} retrieves a document-specific ICL example using BM25~\citep{robertson1995okapi} from the pool of previously human-corrected documents. The closest match record is used as a 1-shot prototype~\citep{ghosh-etal-2024-toward}. This dynamic ICL selection ensures that the LLM receives the most relevant example, improving extraction fidelity. This is a key distinction from other systems (e.g., Elicit and SciSpace).



\subsection{Verification Support module}\label{ssec:support}
To allow experts to verify the generated content, \textsc{SciLire} provides several support tools:

\paragraph{Provenance checking.} We implement a cell-level hallucination checker that compares generated content against the source document using fuzzy string matching \texttt{fuzzywuzzy} library\footnote{\url{https://pypi.org/project/fuzzywuzzy}}. The system visualises match strength as a graded signal, enabling users to build trust in generated answers by identifying aligned and unaligned answers. 

\paragraph{On-demand explanations and insights.} \textsc{SciLire} provides justifications of AI answers by aligning generated text to the source material, implemented as a search function that retrieves the top three supporting paragraphs from the original document using BM25.  Aligned words are shown highlighted in bold to the user to assist data verification. We also provide the detected figures and tables from the PDF as additional resources.  

\textsc{SciLire} also facilitates user requests for LLM-based explanations that identify source paragraphs relevant to a generated answer  (Appendix~\ref{sec:appendix_prompts}).\footnote{Currently, \textsc{SciLire} uses \texttt{GPT-4o} as the model, but any model can be used.}  Together, these mechanisms support the user's verification and curation tasks. 


\section{Experiments}
\label{sec:benchmarking}

Here we report on intrinsic benchmarking experiments to answer the following research questions: \textbf{(RQ1)} Can dynamic sampling (with data from a HAT-DC iterative workflow) lead to improved data extraction? \textbf{(RQ2)}
How does \textsc{SciLire} (and the HAT-DC approach) compare to existing tools providing dataset generation capabilities?

\subsection{Evaluation Framework}\label{sec:evaluation}
\subsubsection{Data and Metrics}
To evaluate the dataset generation capabilities of \textsc{SciLire}, we conduct experiments on 18 datasets from five scientific domains, covering varying levels of granularity in data curation (Appendix~\ref{ssec:appendix_dataset}). We report on a primary evaluation metric which focuses on the very strict \textit{record-level} $F_1$ evaluation rather than cell-level evaluation, as typically scientists are compiling a set of scientific findings that comprise several dependent fields.\footnote{\url{https://github.com/bolucunecva/table_generation}; See Appendix~\ref{ssec:appendix_evaluation_metrics} for detail of the evaluation metrics considered.}

\begin{table}[]
 \resizebox{\columnwidth}{!}
{
\begin{tabular}{r|r|r| r|r|r}\toprule

\bf Dataset  & \multicolumn{1}{c|}{\textbf{0-shot}} & \multicolumn{1}{c|}{\textbf{ICL-10}} & \multicolumn{1}{c|}{\textbf{ICL-50}} & \multicolumn{1}{c|}{\textbf{ICL-100}} & \multicolumn{1}{c}{\textbf{ICL-all}}\\
\midrule
TDMS &  10.14 & 19.01&  23.01 & 24.54 & \textbf{25.02} \\
SciREX & 3.66 & 13.51 & 15.49 & 15.71 & \textbf{18.27} \\
MPEA &  29.23 & 32.52 & 30.39 & \textbf{30.81} & 30.64 \\
 Diffusion & 17.52 & \textbf{17.99} & 17.59 & -- & 17.20 \\
 YSHEAY & 5.34 & 7.87 & \textbf{8.30} & 8.03 & 7.93 \\
 CCRMG &1.82 & 2.48 & -- &   -- & \textbf{2.89}  \\
 Doping & 7.55 & 13.23 & \textbf{14.65} & -- & 12.95 \\
 MMD & 0.78 & -- & -- & -- & \textbf{8.54} \\
 MRL & 1.80 & 1.99 & \textbf{2.04} & -- & 1.82 \\
PNCExtract & 31.14 & 40.21 & 42.46 & \textbf{45.04} & 43.59 \\
 PolyIE & 14.29 & 18.58 & 18.64  & -- & 18.34 \\
BRENDA\_enzyme & 26.58 & \textbf{36.81} & 34.33 & 36.14 & 36.59 \\
 BRENDA\_ribozyme &  11.74 & 17.84 & \textbf{19.12} & 18.69 & 18.48 \\
OPE &  22.33 & \textbf{28.91} & 28.17 & 23.80 & 22.12\\
PPE & 51.67 & \textbf{67.83} & 64.76 & 64.60 & 64.60 \\
 SE & 36.6 & 42.71 & \textbf{47.00} & -- & 46.78 \\
 AE & 15.17 & \textbf{17.90} & 16.14  & -- &  11.53 \\
SuperMat & 5.66  & \textbf{19.31}  & 16.58 & 16.83 & 17.14 \\
\midrule
AVG. & 16.28 & 23.45 & 24.92 & \textbf{28.42} & 22.47 \\
\bottomrule
\end{tabular}
}
\caption{F$_1$ results across datasets.  LLM: GPT-5. 
F$_1$ reported with 0–100 scale; best score is \textbf{boldfaced}. For the full table, see Table~\ref{tab:results_icl_gpt5}.
}
\label{tab:results_icl_F1only}
\end{table}

\subsection{RQ1. Evaluating a HAT-DC Approach} 
Using the benchmark data, we evaluate \textsc{SciLire}'s effectiveness in supporting data curation through HAT by simulating user corrections over multiple scientific domains. With a random sample of papers as an initial data pool, we use the associated human-authored ground truth data from that pool as a stand-in for the corrected records (see Appendix~\ref{ssec:workflow_mimic}).  We then apply dynamic sampling from that pool to create ICL prompts for use with GPT-5.\footnote{GPT-5 was found to be the best performing LLM overall. See Appendix~\ref{ssec:appendix_models} for performance of each tested LLM.} 
Table~\ref{tab:results_icl_F1only} reports the summary results.\footnote{Here we report on 1-shot ICL, guided by an engineering trade-off, given finite context, to balance between ICL examples and the flexibility of the system to accept an arbitrarily long list of columns.}

In line with prior work~\citep{jiang-etal-2024-tkgt}, we observe that generating accurate records is a hard task.  The best reported $F_1$ score is 67.83 (PPE dataset).  The best averaged $F_1$ score was just 28.42, highlighting the complexity of matching full records.  Despite this, we see that the results support the HAT-DC approach: (1) all ICL variants are better than the zero-shot performance, 
and (2) using a sample pool of $n=100$ generally leads to the best performance with marginal gain or even performance degradation beyond that, as the increasing pool size tends to introduce redundancy rather than informative diversity, since dynamic sampling here results in the inclusion of samples that are highly similar.

\begin{table}
\resizebox{\columnwidth}{!}
{
\begin{tabular}{r|r|r|r|r|r|r}\toprule
&  \multicolumn{2}{c|}{\textbf{SciSpace}} & \multicolumn{2}{c|}{\textbf{Elicit}} & \multicolumn{2}{c}{\textbf{ \textsc{SciLire}}} \\\midrule
\bf Dataset &  \multicolumn{1}{c|}{\textbf{0-shot}} &  \multicolumn{1}{c|}{\textbf{ICL-S}} &  \multicolumn{1}{c|}{\textbf{0-shot}} &  \multicolumn{1}{c|}{\textbf{ICL-S}} &  \multicolumn{1}{c|}{\textbf{0-shot}} &  \multicolumn{1}{c}{\textbf{ICL-D}} \\\midrule
TDMS &   0.0 & 0.0  & 0.0 & 3.13 & 3.97 & \textbf{11.76}\\
SciREX &  0.0 & 0.0 & 1.08 & 6.49 & 2.98 & \textbf{18.22}\\
MPEA &13.26 & 13.26 & 0.0 & 0.0 & 40.67 & \textbf{42.27}\\
 Diffusion & 0.65 & 0.65 & 0.06  & 0.53 & 6.80 & \textbf{8.71} \\
 YSHEAY &0.0 &  0.0 & 2.22 & \textbf{13.33} & 3.29 & 5.54\\
 CCRMG  & 0.0 & 0.0 &  0.0 & \textbf{22.22} & 1.80 & 2.67\\
 Doping &   0.0 &  0.0 & 3.6 & 9.01 &  5.41 & \textbf{12.12}\\
 MMD &  0.0 & 0.0 & 0.0 &  0.26 &  0.78 & \textbf{8.65} \\
 MRL & 0.13 &  0.13 &  0.0 & 0.58 & \textbf{1.75} & 1.57 \\
PNCExtract & 2.56 & 2.56 &  5.13 & 5.86 & 29.69 & \textbf{34.96}\\
 PolyIE &   0.0 &  0.0 &  0.0 & 0.74 & 12.05 & 1\textbf{8.58}\\
BRENDA\_enzyme &  0.05 & 0.33 & 0.42 & 1.35 & 34.34& \textbf{47.44} \\
 BRENDA\_ribozyme & 0.0 & 0.0 & 1.96 & 4.34 &  26.03 & \textbf{30.95}\\
OPE &  16.86 & \textbf{20.69} & 10.73 & 16.86 & 19.37 & 16.25 \\ 
PPE &  5.41 & 0.0 &  1.80 & 12.61 & 48.83 & \textbf{62.69} \\
 SE & 0.0 &  0.0 & 0.78 &  0.78 &  34.12 & \textbf{45.25}\\
 AE &  0.0 & 0.0 & 0.0 & 0.0 & \textbf{21.23} & 15.31\\
SuperMat & 4.67 & 0.0 & 0.67 & 1.0 & 11.58 & \textbf{26.80} \\
\midrule
AVG. & 2.42 & 2.09 & 1.58  & 5.50 &  16.93 & \textbf{22.76}  \\
\bottomrule
    \end{tabular}
    }
    \caption{F$_1$ results across datasets comparing \textsc{SciLire} with other data generation tools.
    F$_1$ reported with 0–100 scale; best score is \textbf{boldfaced}. 
    \textsc{SciLire} results are based on GPT-5.  Abbreviations: ICL-S: ICL static, ICL-D: ICL Dynamic ($n$=all). For the full table, see Table~\ref{tab:comparison_tools}. Results are shown for 10 randomly selected PDFs. 
    }
    \label{tab:comparison_tools_F1}
\end{table}

\subsection{RQ2. Related Commercial Software}
As outlined in the Section~\ref{sec:related}, SciSpace and Elicit are comparable to \textsc{SciLire} as they also allow users to curate datasets. 
We evaluate the standard version of these commercial tools and our 1-shot usage of these tools, where we co-opt the column header input textbox in the UI to provide a statically chosen prototype example.\footnote{We use the Extract Data tools of SciSpace and Elicit for comparison.} 
We use one randomly selected static sample per dataset as the static example. 

Elicit, SciSpace, and \textsc{SciLire} can process a different number of PDFs, with Elicit providing the lower bound on PDF uploads.  Here, we selected a random sample of 10 PDFs from each dataset to create a data subsample (in total 180 PDFs) used with each tool. 
The results are given in Table~\ref{tab:comparison_tools_F1}.\footnote{A full comparison between SciSpace and \textsc{SciLire} across all datasets is given in Table~\ref{tab:all_comparison}.}
SciSpace outperforms Elicit in a zero-shot setting. However, Elicit is able to better utilise static ICL.  Ultimately, \textsc{SciLire} consistently outperforms both tools, due to its dynamic sampling (ICL) capability.  This highlights the benefits of the key differentiator of \textsc{SciLire}: the adoption of the HAT-DC approach over a single-pass AI approach. 


\section{Case studies}

To complement the intrinsic benchmarking results, we provide an overview of real usage.
We report on four case studies from different scientific domains, with each study involving one or more researchers engaged in their own data curation tasks.






%
In each case study, domain expert scientists worked with the system development team to document task goals and evaluate progress. 
Variables such as the number of documents per pilot phase iteration were determined by the experts, who were able to seek advice from the development team.
%
Users were free to judge when to end the pilot phase and proceed to the batch phase.

\subsection{Scenarios}
\paragraph{Agriculture}
The scientist was interested in extracting a dataset of reported plant-pest interactions (e.g., plant taxon, insect taxon).  This research is ongoing with a desire to extend the extraction to thousands of articles. Here, the scientist performed multiple rounds in the pilot phase (10-20 articles), and the batch phase included 100 articles.  The results of the batch phase were then manually verified by the scientist.   

\paragraph{Environmental Studies}
The scientist was interested in performing a meta-analysis, collating environmental datasets published in academic publications and grey literature. The aim was to survey the field to determine a standard data schema and then to release a harmonised version of the amalgamated data.  
In the pilot phase of the \textsc{SciLire} workflow, the user conducted 4 rounds of validation on 5 documents each, before scaling up to perform the batch phase on the remainder of the dataset (approx. 200 documents). Because the results were intended for publication, high accuracy through human validation of the full table was required.  


\paragraph{Biochemistry}
This multi-user team was interested in extracting and classifying bioactivity information on various plant compounds. Their goal in using \textsc{SciLire} was to find the ``needles in the haystack'': the rare documents describing specific types of bioactivity. Because such information was scarce, the case study involved validating a mostly sparse table. The users performed the pilot phase through 3 iterations with 18 documents, before electing to perform the batch phase workflow on an additional 81 documents.   

\paragraph{Medical manufacturing}
In this case study, the scientists wished to extract and curate a small dataset (approx. 30 documents) reporting on drug trials. The extracted table included factors related to the drug formulation, experimental design and application area.

\begin{table}[]
 \resizebox{\columnwidth}{!}
{
    \centering
    \begin{tabular}{c c c c c c c}
    \toprule
    \bf Domain & \bf \# Docs & \bf \# Records & \bf \begin{tabular}[c]{@{}c@{}}Edits\\(0-shot)\end{tabular}  & \bf \begin{tabular}[c]{@{}c@{}}Edits\\(ICL)\end{tabular} & \bf \begin{tabular}[c]{@{}c@{}}Time\\(0-shot)\end{tabular} & \bf \begin{tabular}[c]{@{}c@{}}Time\\(ICL)\end{tabular}  \\
    \midrule
    Agri. & 42 & 96 & 30 &  20  & 6  & 7  \\
    Env. St. & 20 & 20 & 35 &  3  & 13  & 3  \\
    Biochem. & 18 & 56 & 7 &  5  & 11  & 3  \\
    Med Man. & 15 & 15 & 12 &  25  & 6  & 9  \\
    \bottomrule
    \end{tabular}
    }
    \caption{User validation behaviour across case studies at the initial (zero-shot) and final pilot phases (with ICL). Edits: Percentage of values edited by the user. Time: Average time (minutes) curating/correcting each PDF. 
    }
    \label{tab:scenario_statistics}
\end{table}

\subsection{Overview of User Interactions}

\begin{figure}[t]
    \centering
    \includegraphics[width=0.9\columnwidth]{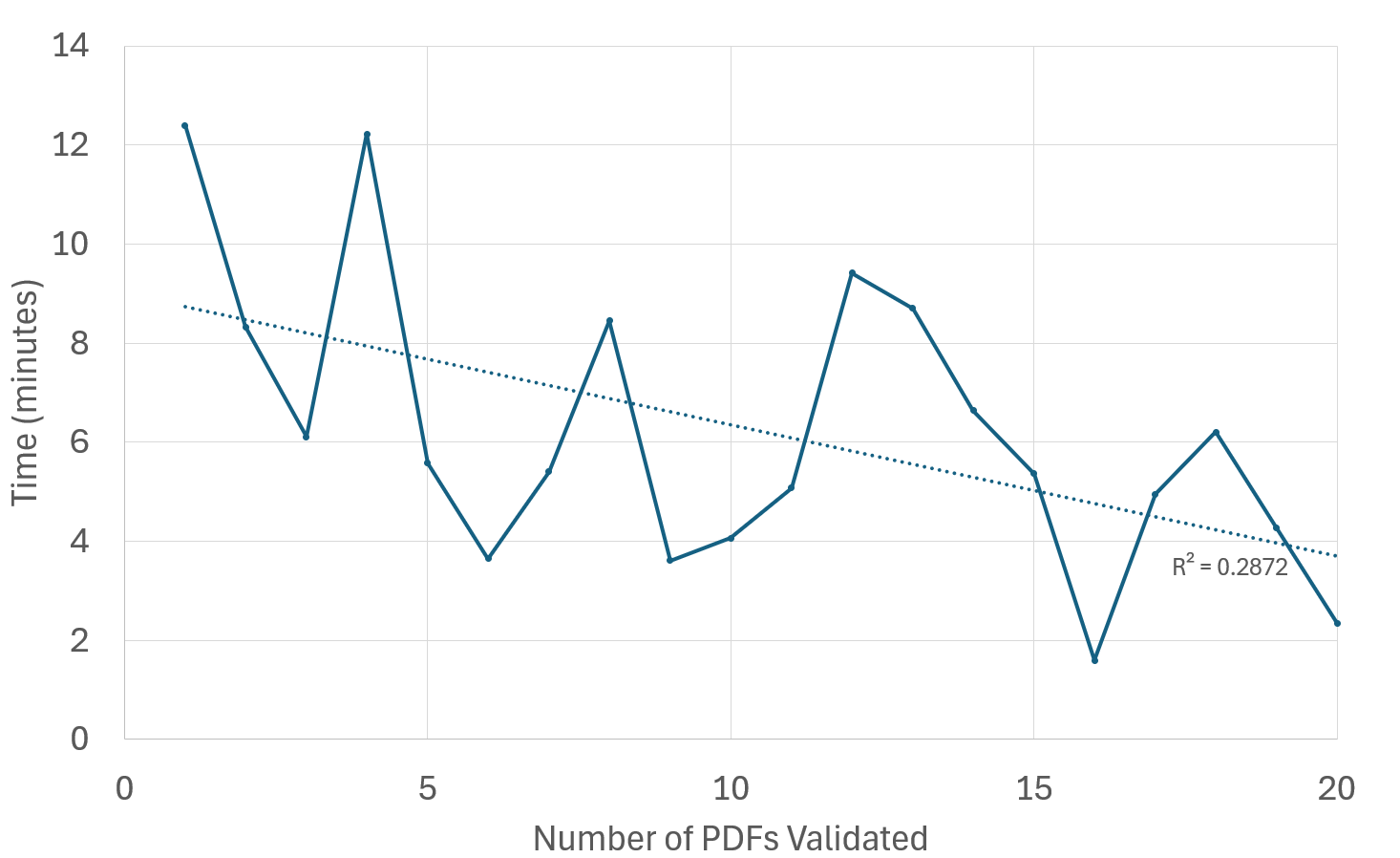}
    \caption{Average time spent validating data from the first 20 papers across the early adopter user cases.}
    \label{fig:time_per_pdf}
\end{figure}

Table~\ref{tab:scenario_statistics} presents an overview of four case studies (e.g., number of documents checked; number of records completed). These statistics confirm that a reference article can yield more than one record.  In general, more edits are made in the initial pilot phase (a zero-shot setting),
and in 3 of the 4 cases, in the dynamic sampling (ICL) setting,
both the number of required edits and the time to complete the task drop.
%
%
We see further evidence of the benefits of the HAT-DC approach in Figure~\ref{fig:time_per_pdf}, which shows the average time spent to validate data for the first 20 PDFs, averaged over the case studies.  While there is a high degree of variability (std.err.: 2.5 mins/paper), we do see a significantly decreasing trend in time ($p<0.025$), indicating a decreasing validation workload as curation interaction increases.

\section{Discussion and User Feedback}
While quantitative benchmarks validate the performance of \textsc{SciLire}, qualitative analysis can provide a deeper understanding of its practical strengths. This section presents user feedback that shows the real-world impact of the system. 

\paragraph{Verifying the need for data verification tools}

As in earlier findings (e.g.,~\citet{Naddaf2025}), interview data with users indicated that they were not prepared to trust AI-generated results in a fully automated (zero-shot) setting. That is, they wanted to inspect and review the content, with the ability to correct the results. This is consistent with our interaction analysis  (see Appendix~\ref{sec:appendix_interaction}), where we see that experts will prefer to verify the data before accepting or rejecting it.  
We observed that how the user performed this review was idiosyncratic: some preferred 
to check the source PDFs, while others preferred the supporting paragraphs.
\textit{This validates our design decision to include tools to facilitate the curation of AI-generated data.}

\paragraph{Opportunities in efficiency and scale}

Users engaging with \textsc{SciLire} saw opportunities on two fronts: (a) \textbf{Scale}, the ability to generate datasets at a scale that would not be feasible manually; and (b) \textbf{Efficiency}, the ability to create datasets with less manual effort. Scaling emerged as the most frequently reported need from users. For example, the agriculture user reported that, given the need to process over 6,000 documents, they could not attempt the task manually. Another noted that for historical datasets, \textsc{SciLire} allowed revisiting documents to add columns on experimentation context, a task that would otherwise not be feasible manually, given the dataset size. Users noted that the dataset compilation task would normally be performed by a team of researchers; with \textsc{SciLire}, it could now be managed by one researcher with a modest budget to cover the additional computational costs.

Given the fallibility of AI, users reported that they preferred checking pre-populated fields to manually populating the table from scratch. 
They felt that having a starting point from which to start the validation process increased their efficiency. 
%
This efficiency was recognised by users as time savings.  For example, one user estimated that if they were to redo a recent manual systematic review with \textsc{SciLire}, they might perform the task twice as fast. Another estimated that \textsc{SciLire} reduced the time required from 2-3 months to 1 month. 
%
%
%
\textit{This validates the design decision to consider data curation AI tools leading to productivity benefits.}

\section{Conclusion}
We presented \textsc{SciLire}, a Human-AI Teaming (HAT) system, where AI capabilities and human expert judgements work together to enable effective dataset creation from scientific literature.  Our intrinsic benchmarking demonstrates dynamic samples for few-shot learning (from an iterative workflow), guides, and improves AI results.  These outcomes mirror findings in our user studies, which reveal that the \textsc{SciLire} and the HAT approach lead to new opportunities for scientists.
With an AI-enabled data creation and curation workflow, users can work at scales that would be infeasible without AI support, while gaining efficiency in overseeing and validating the AI-generated results.

As future work, we plan to extend \textsc{SciLire} with additional analysis capabilities, allowing users to transform curated datasets into actionable insights by facilitating the discovery of trends and patterns.

\section*{Acknowledgments}
This work is supported by CSIRO as part of the AI for Missions Program (\url{https://research.csiro.au/ai4m/}). We acknowledge the CSIRO Language Technology team and Samuel Walker, a former CSIRO employee (UX Engineer), for their contributions to the \textsc{SciLire}.

\section*{Ethical Consideration}
The public datasets used in quantitative experiments are adopted from existing repositories. Therefore, we do not foresee any serious or harmful issues related to their content. We have collected documents with the given identifiers (e.g., DOI, PubMed ID). 

Collection of user data was approved by the CSIRO Social and Interdisciplinary Science Human Research Ethics Committee (090/23), and participants whose data are presented here provided informed written consent.



\bibliographystyle{acl_natbib}
\bibliography{custom}

\appendix
\section{Prompts}\label{sec:appendix_prompts}
The prompt used in the record generation module:

\begin{myverbatim}
Please, extract ATTRIBUTE_1, ATTRIBUTE_2,
..., ATTRIBUTE_n from the given article.

For the extracted information, you MUST 
respond in a list of JSON dictionaries 
structure with the given 
Dictionary Key Mapping.

[Dictionary Key Mapping in your response]
{
ATTRIBUTE_1: (example: VALUE_1),
ATTRIBUTE_2: (example: VALUE_2),
...
ATTRIBUTE_n: (example: ATTRIBUTE_n)
}

[Given Article Start]
ARTICLE CONTENT
[Given Article End]
\end{myverbatim}

The prompts used in on-demand explanations by LLMs:

\begin{myverbatim}
Please find the relevant paragraph that 
shows that the ATTRIBUTE is VALUE 
from given article.

[Given Article Start]
ARTICLE CONTENT
[Given Article End]
 \end{myverbatim}

\section{Table \& Figure Extraction Module}\label{sec:appendix_table_recognition}

\begin{figure}[th]
    \centering
    \includegraphics[width=0.8\columnwidth]{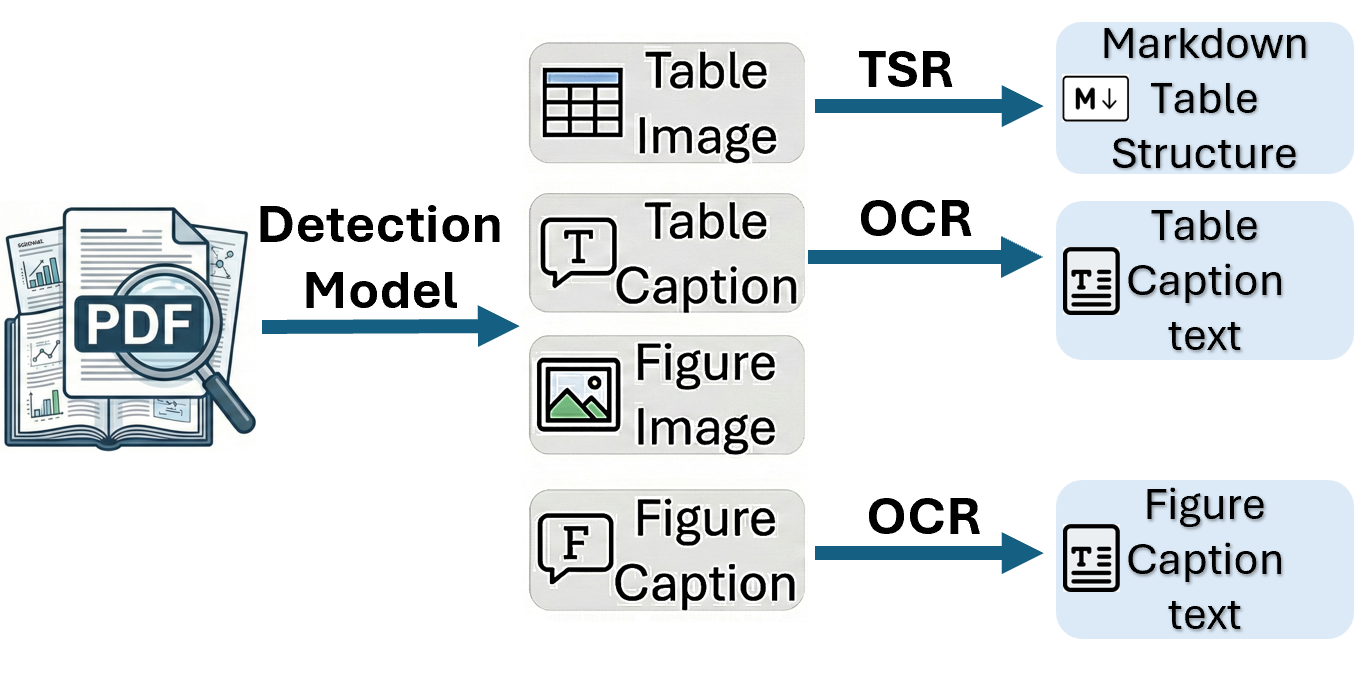}
    \caption{The Table \& Figure Extraction module.}
    \label{fig:tr_pipeline}
\end{figure}

Our pipeline (Figure~\ref{fig:tr_pipeline}) consists of two main stages: (1) table, figure, and caption detection, and (2) table structure recognition (TSR). 

\subsection{Stage I: Table, figure, caption detection}

\paragraph{Training.} The detection model architecture is based on \texttt{Cascade R-CNN}~\cite{cai2018cascade}. We fine-tune the pre-trained (on COCO~\cite{lin2014microsoft} dataset) Cascade R-CNN  using the SCI-3000 dataset~\cite{sci3000}\footnote{The original SCI-3000 dataset uses a single caption label for both tables and figures. For our task, we re-annotate the captions to separate table captions from figure captions to provide finer granularity. The resulting dataset contains four labels: table, table caption, figure, and figure caption.} to detect tables, figures, and their captions from PDF page images. All the implementations are based on Detectron2\footnote{\url{https://github.com/facebookresearch/detectron2}}. The results are given in Table~\ref{tab:detection}.


\begin{table}[h]
\resizebox{\columnwidth}{!}
{
  \begin{tabular}{lccc}
    \toprule
    \textbf{Model} & \textbf{mAP@.5} & \textbf{mAP@.75} & \textbf{mAP} \\
    \midrule
    Fast R-CNN~\citep{girshick2015fast}  & 96.18 & 94.20 & 87.36 \\
    Mask R-CNN~\citep{he2017mask}  & 96.21 & 94.39 & 87.57 \\
    Cascade R-CNN & \textbf{97.12} & \textbf{95.40} & \textbf{90.55} \\
    \bottomrule
  \end{tabular}
  }
  \caption{Evaluation results comparing Cascade R-CNN with baseline approaches. The best results are \textbf{boldfaced}.}
  \label{tab:detection}
\end{table}

\begin{table}[!h]
\resizebox{\columnwidth}{!}
{
  \begin{tabular}{lcc}
    \toprule
    \textbf{Model} & \textbf{TEDS} & \textbf{\textbf{mAP@.5}} \\
    \midrule
    UniTable~\cite{peng2024unitable} & 95.23 & 96.23 \\
    UniTable (finetuned on SciTSR) & \textbf{97.98} & \textbf{96.98} \\
    Qwen2.5-VL-72B~\cite{qwen2025qwen25technicalreport} & 82.72 & --- \\
    \bottomrule
  \end{tabular}
  }
  \caption{Evaluation results comparing UniTable with and without finetuning. The best results are \textbf{boldfaced}.}
  \label{tab:tsr}
\end{table}

\begin{table*}[th]
\resizebox{\textwidth}{!}
{
\begin{tabular}{rc rrrrrr}\toprule
      \bf Domain  & \bf Dataset   & \bf \# Documents & \bf Avg. Pages & \bf \# Records & \bf Records / Document &  \bf Schema Size & \bf Reference   \\\midrule
       \multirow{2}{*}{Machine Learning} & TDMS &  332 & 10.52 & 904 & 2.72 & 4 & \citet{hou2019identification}\\
        & SciREX &  372 & 11.83 & 1,897  & 5.34  & 5 & \citet{jain2020scirex}\\ \midrule
        \multirow{9}{*}{Material Science} & MPEA &  264 & 8.70 & 1,544 & 5.85 & 17 & \citet{borg2020expanded}  \\
       & Diffusion  &  93 & 14.02 &  3,533 & 37.99 & 18 & \citet{zhang2010diffusion}\\ 
        & YSHEAY &  219 & 15.65 & 837  &  3.82 & 3 & \citet{polak2024flexible} \\ 
         &CCRMG &  24 & 11.96  &  43 & 1.79 &  3 & \citet{polak2024flexible}\\
       & Doping &   66 & 6.66 &  544 & 8.24 & 3 & \citet{dunn2022structured}\\ 
       & MMD &   3 & 6.00& 140 & 46.67 &  16 & \citet{xie2023darwin}\\ 
        & MRL &   100 & 5.81 &  993 & 9.93 &  19 & \citet{xie2023darwin}\\ 
         & PNCExtract &   155 & 8.92 &  838 & 5.41 & 6 & \citet{khalighinejad2024extracting} \\
       & PolyIE &   76 & 8.58&  2,337 & 30.75 &  3 & \citet{cheung2024polyie}
       \\\midrule 
    \multirow{5}{*}{Chemistry} & BRENDA-enzyme &   155 & 12.45 & 4,210  & 27.16 & 13 & \citet{jiang2025enzyme} \\
    & BRENDA-ribozyme &   163 & 11.05 & 1,756 & 10.77 & 17 & \citet{jiang2025enzyme} \\
       & OPE &  104 & 7.55 &  255 & 2.45 & 7 & \citet{cai2024sciassess}\\ 
       & PPE &  109 & 6.97 &  265 & 2.43 & 6 & \citet{cai2024sciassess}\\ 
        & SE &  96 & 8.01 &2,363 & 24.61 & 3 & \citet{cai2024sciassess}\\\midrule
 Medicine & AE &  40 & 4.30 & 406 & 10.15 &  3  & \citet{cai2024sciassess}\\\midrule
Physics & SuperMat & 142 & 8.55 & 1,301 & 9.16 & 4  &  \citet{doi:10.1080/27660400.2021.1918396}\\
\bottomrule
    \end{tabular}
    }
    \caption{Statistics of datasets across five domains.
}
    \label{tab:dataset}
\end{table*} 

\paragraph{Inference.}
We convert each PDF page into an image. The detection model then identifies tables, figures, and their captions on each page.

\subsection{Stage II: Table structure recognition}

\paragraph{Training.} Table structure recognition architecture is based on \texttt{UniTable}~\cite{peng2024unitable}. We fine-tune the model on the SciTSR dataset~\cite{chi2019complicated}, which contains 15k table images (12k for training and 3k for testing) and their corresponding structure labels obtained from LaTeX source files. We convert the SciTSR data annotation format to HTML to align with the {UniTable} training and TEDS evaluation format. The comparison results between the base model and our fine-tuned model are shown in Table~\ref{tab:tsr}. Since the model predicts HTML-formatted table structures, we convert the predicted HTML into markdown as the final output. 

\paragraph{Inference.} The TSR model reads each detected table image and infers the table structure along with its content. It outputs the table in markdown format, the extracted cell content, and the corresponding caption (extracted by \texttt{Tesseract OCR}~\citep{smith2007overview}).

\section{Experimental Details}\label{sec:appendix_ex_details}
\subsection{Datasets}\label{ssec:appendix_dataset}
For the quantitative experiments (Section~\ref{sec:evaluation}), we use datasets from multiple domains spanning machine learning, materials science, chemistry, medicine, and physics\footnote{You can find more details about datasets (e.g., schema) at \url{https://github.com/bolucunecva/table_generation}}. Together, these datasets provide broad, heterogeneous, and real scientific documents for evaluating \textsc{SciLire}. 

\begin{itemize}[nosep]
    \item \textbf{Machine Learning:} Leaderboard construction task from NLP papers (TDMS, SciREX).
    \item \textbf{Materials Science:} Extraction of compositions, experimental parameters, and material properties from diverse subfields (MPEA, Diffusion, YSHEAY, CCRMG, MRL, Doping, MMD, PolyIE, PNCExtract).
    \item \textbf{Chemistry:} Extraction of molecular structures, reaction properties, and material characteristics (BRENDA, OPE, PPE, SE).
    \item \textbf{Medicine:} Affinity extraction involving molecules, SMILES, and bioassay targets (AE).
    \item \textbf{Physics:} Extraction of superconductor materials and properties (SuperMat).
\end{itemize}

The statistical details of datasets are given in Table~\ref{tab:dataset}.

\subsection{Experimental Settings of HAT-DC Workflow}\label{ssec:workflow_mimic}
Algorithm~\ref{alg:HAT-DL} outlines the HAT-DC-mimicking workflow. This procedure allows us to simulate iterative human-AI interactions and evaluate the benefits of HAT-DC guidance in a controlled, reproducible manner. As a reminder, the user actively selects samples for a sample pool for dynamic sampling in \textsc{SciLire}, rather than random sampling.

\begin{algorithm}[t]
\small
\caption{Mimic HAT-DC Workflow}
\SetKwFunction{BM}{BM25Rank}
\SetKwFunction{Pred}{Prediction}
\SetKwFunction{RS}{RandomSample}

\KwIn{Dataset $D$; pool size $k$; samples $m$; schema $H$; LLM $LLM$}
\KwOut{Table $\mathsf{T}$}

$\mathsf{T} \gets [\,]$\;
$S \gets$ samples of $D$; $N \gets |S|$\;

\For{$t \gets 1$ \KwTo $N$}{
    $\texttt{test} \gets S[t]$\;
    $\texttt{trainCandidates} \gets S \setminus \{\texttt{test}\}$\;
    $\texttt{pool} \gets \RS(\texttt{trainCandidates}, k)$\;
    $\texttt{ranked} \gets \BM(\texttt{test}, \texttt{pool})$\;
    $\texttt{sample} \gets \texttt{ranked}[1{:}m]$\;
    $\texttt{records} \gets \Pred(LLM, \texttt{test}, \texttt{sample}, H)$\;
    append $\langle\texttt{records}\rangle$ to $\mathsf{T}$\;
}

\Return{$\mathsf{T}$}
\label{alg:HAT-DL}
\end{algorithm}

\subsection{Evaluation metrics}\label{ssec:appendix_evaluation_metrics}

We adapt table-generation metrics for record-level evaluation~\citep{ghosh-etal-2024-toward, khalighinejad2024extracting, cheung2024polyie, feng2024sciknoweval, jiang2025enzyme}. Precision, Recall, and F$_1$ are computed with, where a cell counts as correct only if it exactly matches the aligned reference. ChrF is also reported in the record-aligned setting, measuring character n-gram overlap to capture partial matches and minor differences. 

\subsection{Models}\label{ssec:appendix_models}
The details of the models used in the record generation module of \textsc{SciLire} (Section~\ref{ssec:record_generation}) are given in Table~\ref{tab:models}.

\begin{table}[th]
\resizebox{\columnwidth}{!}
{
\begin{tabular}{r rrrr}\toprule

\bf Model & \bf \# of Par. & \bf Context Length & \bf Open-Source & \bf Family  \\
\midrule
GPT-OSS:20b & 20B & 128K & \cmark & OpenAI \\
GPT-OSS:120b & 120B & 128K & \cmark & OpenAI \\
Gemma3:1B& 1B & 32K & \cmark & Google \\
Gemma3:4B & 4B & 128K & \cmark & Google \\
Gemma3:12B & 12B & 128K & \cmark & Google \\
Gemma3:27B & 27B & 128K & \cmark & Google \\
Qwen3:0.6B & 0.6B & 40K & \cmark & Qwen \\ 
Qwen3:4B & 4B & 256K & \cmark & Qwen \\ 
Qwen3:14B & 14B & 40K & \cmark & Qwen \\ 
Qwen3:32B & 32B & 40K & \cmark & Qwen \\ 
Phi-4 & 14B & 16K & \cmark & Microsoft \\
DeepSeek-R1-Llama:8B & 8B & 128K & \cmark & DeepSeek \\
DeepSeek-R1-Llama:70B & 70B & 128K & \cmark & DeepSeek \\
DeepSeek-R1-Qwen3:14B & 14B & 128K & \cmark & DeepSeek \\
DeepSeek-R1-Qwen3:32B & 32B & 128K & \cmark & DeepSeek \\
GPT-5 & ? & 400K & \xmark & OpenAI \\
\bottomrule
\end{tabular}
}
\caption{Details of models used in record generation module (Section~\ref{ssec:record_generation}) of \textsc{SciLire}.}
\label{tab:models}
\end{table}

\subsection{All Experimental Results}\label{ssec:appendix_all_experiments}

The overall experimental results across all datasets and models are summarised in Table~\ref{tab:results_summary}. Table~\ref{tab:results_icl_gpt5} the full per-dataset results using GPT-5 as LLM.

Detailed tool-level comparisons are reported in Table~\ref{tab:comparison_tools}, and Table~\ref{tab:all_comparison} presents a comprehensive comparison between \textsc{SciLire} and SciSpace across all datasets using the complete document collections.

\begin{table*}[]
 \resizebox{\textwidth}{!}
{
\begin{tabular}{r| rrrr | rrrr | r | rr | rrrr | r}\toprule

\bf Dataset  & \begin{turn}{90} Gemma3:1B\end{turn} &
\begin{turn}{90} Gemma3:4B\end{turn} & \begin{turn}{90} Gemma3:12B\end{turn}&
\begin{turn}{90} Gemma3:27B\end{turn}  & \begin{turn}{90} Qwen3:0.6B\end{turn} & \begin{turn}{90}Qwen3:4B\end{turn} & \begin{turn}{90} Qwen3:14B\end{turn} & \begin{turn}{90}Qwen3:32B\end{turn} &
\begin{turn}{90}Phi-4\end{turn} & \begin{turn}{90}GPT-OSS:20B\end{turn} & \begin{turn}{90}GPT-OSS:120B\end{turn} &
\begin{turn}{90}Deepseek-R1-Llama:8B\end{turn} & \begin{turn}{90}Deepseek-R1-Llama:70B\end{turn} & \begin{turn}{90}Deepseek-R1-Qwen3:14B\end{turn} & \begin{turn}{90}Deepseek-R1-Qwen3:32B\end{turn} & \begin{turn}{90}GPT-5\end{turn}\\
\midrule
TDMS & 12.20 & 12.13 & 18.80 & 21.90 & 17.13 & 25.09 & 17.98 & 21.91 & 16.10 & 22.10 & 23.02 &  21.19 & 22.53 & \textbf{26.04} & 23.13 & 25.02\\
SciREX & 4.10 & 3.97 & 14.45 & 15.08 & 7.82 & 15.69 & 17.06 & 16.53 & 10.56 & 18.12  & 17.65 & 6.26 & 08.45 & 15.99 & 14.56& \textbf{18.27} \\
MPEA & 9.29 & 10.45 & 13.45 & 18.36 & 8.43 & 15.23 & 18.43 & 17.56 & 13.54 & 19.45 & 16.34 & 7.21 & 14.56 & 12.42 & 13.21& \textbf{30.64} \\
Diffusion & 0.45 & 4.20 & 5.45 & 8.36 & 0.68 & 6.34 & 9.12& 12.25 & 3.12 & 14.21 &  16.34 & 7.10 & 9.21 & 12.10 & 12.46 &  \textbf{17.20} \\
YSHEAY & 9.23 & 10.13 & 12.21 & 13.10 & 17.35 & 14.25 & 15.99 & \textbf{16.48} & 11.24 &  16.23 &14.56 & 12.48 & 13.21 & 12.02 & 10.45 & 7.93\\
CCRMG &  5.85 & 5.23 & 5.71 & 3.37 & 4.96 & 5.01 & 5.89 & \textbf{9.03} & 8.91 & 6.10 & 7.03 & 2.24 & 3.87 & 5.02 & 4.65 &2.89 \\
Doping & 6.44 & 11.99 & 13.70 & 15.82 & 3.68 & 11.83 & 14.78 & 15.20 & 13.15 & 15.33 & \textbf{16.45} & 9.08 & 10.21 & 14.04 & 13.67 & 12.95\\
MMD & 1.22 & 3.15 & 4.7 & 3.96 & 1.26 & 1.30 & 11.05 & 5.92 & 4.23& \textbf{11.08} & 10.98 & 1.39 & 8.45 & 2.48 & 6.10& 8.54\\
MRL & 0.23 & 0.45 & 0.59 & 1.23 & 0.67  & 1.34 & 1.97 &  \textbf{1.95} & 0.45 & 1.78 & 1.82 & 0.56&1.10 & 1.13 & 1.12& 1.82\\
PNCExtract & 12.66 & 29.29 & 34.30 & 35.78 & 20.82 & 21.07 & 42.49 & 27.86 & 19.92 &  26.12 & 29.48 & 20.73 & 32.87 & 30.35 & 28.23& \textbf{43.59} \\
PolyIE & 3.54 & 17.04 & 21.38 & 23.69 & 6.56 & 16.47 & 20.60 & 19.05 & 17.25 & \textbf{21.10} & 20.03 & 15.12 & 14.32 & 17.51 & 16.18 & 18.34\\
BRENDA\_enzyme & 3.08 & 17.04 & 16.85 & 22.97 & 3.11 & 5.44 & 24.07 & 30.83 & 15.12 & 31.27 & 32.33& 7.31 & 13.23 & 14.45&  13.45& \textbf{36.59} \\
BRENDA\_ribozyme & 1.16 & 3.80 & 6.05 & 7.74 & 2.05 & 2.81 & 6.47 & 14.98 & 4.36 &  15.98 & 14.65  & 5.82 & 6.23 & 5.81 & 8.10 & \textbf{18.48} \\
OPE & 5.34 & 9.23 & 12.27 & 15.20 & 13.21 & 16.05 & 17.54 & 16.89 & 9.24 & 20.13 & 21.05 & 15.32 & 14.23 & 13.45 & 17.89 & \textbf{22.12} \\
PPE & 16.38 & 41.16 & 45.65 & 60.48 &  50.21 & 54.62 & 63.43 & \textbf{66.80} & 55.52 &  66.10  & 65.86 & 34.63 & 54.19 & 56.86 &  55.10& 64.60\\
SE & 4.35 & 24.86 & 29.13 & 31.06 & 9.36 & 23.23 & 43.87 & 43.42 & 32.73 & 45.10 & 44.61 & 17.50 & 22.10 & 30.75 & 31.20 & \textbf{46.78} \\
AE & 1.92 & 8.0 & 19.19 & 21.75 & 2.97 & 21.28 & 23.20 & \textbf{25.89} & 12.54 & 24.45  & 23.76 & 19.22 &  20.10 & 19.63 &  18.65 & 11.53  \\
SuperMat & 12.74 & 22.95 & 28.40 & 23.51 & 10.47 & 11.55 & 13.48 & 13.80 & 19.53 & \textbf{21.0} & 20.14 & 8.75 & 9.56 & 13.41 & 14.32 & 17.14 \\
\midrule
AVG. & 6.12 & 13.06 & 16.79 & 19.07 & 10.04 & 14.92 &  20.41 & 20.91 & 14.86 & 21.98 & 22.01 & 11.77 &  15.47 & 16.86 & 16.80 &  \textbf{22.47}\\
\bottomrule
\end{tabular}
}
\caption{F$_1$ results across datasets for multiple LLMs and their ability to benefit from ICL Dynamic sampling ($n$=all). F$_1$ reported with 0–100 scale; best score is \textbf{boldfaced}. For detailed results, see \url{https://github.com/bolucunecva/scilire}.}
\label{tab:results_summary}
\end{table*}

\begin{table*}[th]
\resizebox{\textwidth}{!}
{
\begin{tabular}{r| rrrr| rrrr| rrrr| rrrr| rrrr}\toprule

 & \multicolumn{4}{c|}{\textbf{Zero-shot}} & \multicolumn{4}{c|}{\textbf{ICL -10}} & \multicolumn{4}{c|}{\textbf{ICL -50}} & \multicolumn{4}{c|}{\textbf{ICL -100}} & \multicolumn{4}{c}{\textbf{ICL -all}}\\
\midrule
\bf Models & \bf P & \bf R & \bf F$_1$ & \bf ChrF &  \bf P & \bf R & \bf F$_1$ & \bf ChrF &  \bf P & \bf R & \bf F$_1$ & \bf ChrF &  \bf P & \bf R & \bf F$_1$ & \bf ChrF &  \bf P & \bf R & \bf F$_1$ & \bf ChrF \\\midrule
TDMS &  6.95 & 18.75 & 10.14 & 11.94 &  13.88 & 30.14& 19.01& 13.88 &  16.47 & 38.2 & 23.01 & 14.70 & 17.68 & 40.13 & 24.54 & 15.11 & 18.12 & 40.38 & \textbf{25.02} & 15.05\\
SciREX & 2.8 & 5.29 & 3.66 & 7.16 & 10.67 & 18.39 & 13.51 & 10.16 & 12.61 & 20.07 & 15.49 & 10.78 & 12.53 & 21.06 & 15.71 & 11.09 & 14.52 & 24.61 & \textbf{18.27} & 11.63 \\
MPEA &  31.12 & 27.56 & 29.23 & 0.74 & 34.59 & 30.68 & 32.52 & 0.91 & 32.9 & 28.24 & 30.39 & 0.77 & 32.45 & 29.32 & \textbf{30.81} & 0.74 & 31.92 & 29.45 & 30.64 & 0.74\\
Diffusion & 33.51 & 11.86 & 17.52 & 1.54 & 32.51 & 12.43 & \textbf{17.99} & 1.58 & 33.13 & 11.98 & 17.59 & 1.6 &  -- & -- & -- & -- & 30.35 & 12.0 & 17.20 & 1.52\\
YSHEAY & 2.94 & 29.51 & 5.34 & 12.32 & 4.45 & 34.41 & 7.87 & 12.96 & 4.69 & 35.96 & \textbf{8.30} & 13.14 & 4.54 & 34.85 & 8.03 & 13.11 & 4.47 & 35.36 & 7.93 & 13.28\\
CCRMG &  0.96 & 18.7 & 1.82 & 11.64 & 1.31 & 23.58 & 2.48 & 13.59 &  -- & -- & -- & -- &   -- & -- & -- & -- &  1.53 & 25.20 & \textbf{2.89} & 12.48 \\
Doping & 20.66 & 4.62 & 7.55 & 10.35 & 23.65 & 9.18 & 13.23 & 14.85 &26.19 & 10.17 & \textbf{14.65} & 14.10 &  -- & -- & -- & -- & 22.32 & 9.12 & 12.95 & 14.57\\
MMD & 2.98 & 0.45 & 0.78 & 1.08 &  -- & -- & -- & -- &   -- & -- & -- & -- &  -- & -- & -- & -- & 37.5 & 4.82 & \textbf{8.54} & 2.12 \\
MRL & 3.54 & 1.20 & 1.80 & 1.12 & 3.95 & 1.33 & 1.99 & 1.14 & 4.22 & 1.34 & \textbf{2.04} & 1.19 &  -- & -- & -- & -- & 4.25 & 1.43 & 1.82 & 2.01\\
PNCExtract & 30.61 & 31.7 & 31.14 & 9.66 & 40.4 & 40.02 & 40.21 & 10.73 & 42.71 & 42.2 & 42.46 & 11.24 & 45.82 & 44.29 & \textbf{45.04} & 11.49 &  44.83 & 42.42 & 43.59 & 11.50\\
PolyIE & 11.43 & 19.04 & 14.29 & 15.67 &15.78 & 22.61 & 18.58 & 15.79 & 16.33 & 21.69 & \textbf{18.64} & 15.75 &  -- & -- & -- & -- &  15.88 & 21.69 & 18.34 & 15.75\\
BRENDA\_enzyme & 36.2 & 21.0 & 26.58 & 4.14 & 52.59 & 28.31 & \textbf{36.81} & 4.94 & 49.72 &  26.21 & 34.33 & 4.77 & 51.14 & 27.94 & 36.14 & 4.91 & 53.08 & 27.92 & 36.59 & 5.06\\
BRENDA\_ribozyme &  14.08 & 10.06 & 11.74 & 2.09 & 21.17 & 15.42 & 17.84 & 2.36 & 22.95 &  16.39 & \textbf{19.12} & 2.51 & 22.36 &  16.05 & 18.69 & 2.49 & 22.47& 15.69 & 18.48 & 2.51\\
OPE & 16.33 & 35.34 & 22.33 & 5.78 & 21.15 & 45.64 & \textbf{28.91} & 7.19 & 20.69 & 44.09 & 28.17 & 7.16 & 17.38 & 37.76 & 23.80 & 5.97 & 16.03 & 35.65 & 22.12 & 5.88\\
PPE & 41.65 & 68.05 & 51.67 & 11.62 & 58.68 & 80.38 & \textbf{67.83} & 14.07 &  55.78 & 78.93 & 65.36 & 13.72 &  53.78 &  81.38 & 64.76 & 13.83 & 53.91 &  80.57 & 64.60 & 13.81 \\
SE & 40.50 &  33.38 & 36.6 & 11.42 & 49.16 & 37.76 & 42.71 & 12.5 & 51.61 &  43.15 & \textbf{47.00} & 12.41&  -- & -- & -- & -- & 51.83 & 42.62 & 46.78 & 12.8\\
AE & 17.25 &  13.53 & 15.17 & 5.12 &20.25 & 16.04 & \textbf{17.90} & 10.03 & 17.77 &  14.79 & 16.14 & 8.87&  -- & -- & -- & -- &  13.56 & 10.03 & 11.53 & 9.34\\
SuperMat & 11.48 & 3.76 & 5.66 & 9.30 &   40.05 & 12.72 & \textbf{19.31} & 11.06 &  33.08 & 11.06 & 16.58 &11.23 & 34.24 & 11.16 & 16.83 &11.32 & 34.63 & 11.39 & 17.14 &11.06\\
\bottomrule
\end{tabular}
}
\caption{Evaluation results across datasets. Cells marked with `--' indicate that the dataset does not have enough train data for evaluation. All scores are reported on a 0–100 scale, with the best F$_1$ score highlighted in \textbf{boldfaced}. LLM: GPT-5.}
\label{tab:results_icl_gpt5}
\end{table*}

\begin{table*}
\resizebox{\textwidth}{!}
{
\begin{tabular}{r| rrrr| rrrr| rrrr | rrrr| rrrr| rrrr}\toprule
&  \multicolumn{8}{c|}{\textbf{SciSpace}} & \multicolumn{8}{c|}{\textbf{Elicit}} & \multicolumn{8}{c}{\textbf{ \textsc{SciLire} best (ICL large)}} \\\midrule
&  \multicolumn{4}{c|}{\textbf{Zero-shot}} &  \multicolumn{4}{c|}{\textbf{ICL (Static)}} &  \multicolumn{4}{c|}{\textbf{Zero-shot}} &  \multicolumn{4}{c|}{\textbf{ICL (Static)}} &  \multicolumn{4}{c|}{\textbf{Zero-shot}} &  \multicolumn{4}{c}{\textbf{ICL (Dynamic)}} \\\midrule
        \bf Dataset & \bf P & \bf R & \bf F$_1$ & \bf ChrF & \bf P & \bf R & \bf F$_1$ & \bf ChrF & \bf P & \bf R & \bf F$_1$ & \bf ChrF & \bf P & \bf R & \bf F$_1$ & \bf ChrF & \bf P & \bf R & \bf F$_1$ & \bf ChrF & \bf P & \bf R & \bf F$_1$ & \bf ChrF \\\midrule
TDMS &  0.0 & 0.0 & 0.0 & 2.97 &  0.0 & 0.0 & 0.0 & 4.36 &  0.0 & 0.0 & 0.0 & 5.43 &  5.0 & 2.27 & 3.13 & 7.70 & 2.4 & 11.36 & 3.97 & 9.42 & 7.94 & 22.73 & \textbf{11.76} & 11.08\\
SciREX & 0.0 & 0.0 & 0.0 & 2.97 & 0.0 & 0.0 & 0.0 & 2.97  & 2.0 & 0.74 & 1.08 & 3.73 & 12.0 & 4.44 & 6.49 & 6.80 &2.09 & 5.19 & 2.98 & 8.09 & 12.43 & 34.07 & \textbf{18.22} & 14.21 \\
MPEA &47.06 & 7.71 & 13.26 & 0.24 &  47.06 & 7.71 & 13.26 & 0.24 & 0.0 & 0.0 & 0.0 & 0.22 &  0.0 & 0.0 & 0.0 & 0.29 & 39.13 & 42.33 & 40.67 & 0.73 & 39.59 & 45.35 & \textbf{42.27} & 0.74\\
Diffusion & 20.99 & 0.33 & 0.65 & 0.35 & 20.99 & 0.33 & 0.65 & 0.35 & 1.85 & 0.03 & 0.06 & 0.55 &  17.28 & 0.27 & 0.53 & 0.96 & 23.29 & 3.98 & 6.80 & 1.41 & 27.51 & 5.17 & \textbf{8.71} & 1.59\\
YSHEAY & 0.0 & 0.0 & 0.0 & 5.12 & 0.0 & 0.0 & 0.0 & 5.12 & 3.33 & 1.67 & 2.22 & 5.35 & 20.0 & 10.0 & \textbf{13.33} & 11.28 & 1.73 & 33.33 & 3.29 & 11.72 & 2.97 & 41.67 & 5.54 & 12.22\\
CCRMG  & 0.0 & 0.0 & 0.0 & 3.83 & 0.0 & 0.0 & 0.0 & 3.92 & 0.0 & 0.0 & 0.0 & 4.12 & 33.33 & 16.67 & \textbf{22.22} & 8.36 & 0.96 & 15.0 & 1.80 & 11.82 & 1.42 & 21.67 & 2.67 & 11.73\\
Doping &  0.0 & 0.0 & 0.0 & 4.47 & 0.0 & 0.0 & 0.0 & 4.04 & 13.33 & 2.08 & 3.60 & 5.05 & 33.33 & 5.21 & 9.01 & 7.61 & 20.0 & 3.12 & 5.41 & 8.32 & 22.22 & 8.33 & \textbf{12.12} & 9.67\\
MMD &  0.0 & 0.0 & 0.0 & 0.29 &   0.0 & 0.0 & 0.0 & 0.25 & 0.0 & 0.0 & 0.0 & 0.30 & 6.25 & 0.13 & 0.26 & 0.46 &  2.98 & 0.45 & 0.78 & 1.08 & 36.18 & 4.91 & \textbf{8.65} & 1.96\\
MRL & 1.05 & 0.07 & 0.13 & 0.24 & 1.05 & 0.07 & 0.13 & 0.23 & 0.0 & 0.0 & 0.0 & 0.40 & 4.74 & 0.31 & 0.58 & 0.64 & 4.66 & 1.07 & \textbf{1.75} & 1.08 & 4.90 & 0.93 & 1.57 & 1.10\\
PNCExtract & 11.67 & 1.44 & 2.56 & 1.56 &  11.67 & 1.44 & 2.56 & 1.56 &  23.33 & 2.88 & 5.13 & 4.43 & 26.67 & 3.29 & 5.86 & 5.05 &  40.43 & 23.46 & 29.69 & 9.64 & 51.19 & 26.54 & \textbf{34.96} & 10.83\\
PolyIE &  0.0 & 0.0 & 0.0 & 4.71 & 0.0 & 0.0 & 0.0 & 4.89 & 0.0 & 0.0 & 0.0 & 4.0 &  13.33 & 0.38 & 0.74 & 7.33 & 10.56 & 14.04 & 12.05 & 13.60 & 17.24 & 20.15 & \textbf{18.58} & 15.11\\
BRENDA\_enzyme &  0.77 & 0.02 & 0.05 & 0.61 & 5.38 & 0.17 & 0.33 & 0.64 & 6.92 & 0.22 & 0.42 & 0.59 &  22.31 & 0.70 & 1.35 & 1.51 & 47.03 & 27.04 & 34.34 & 4.37 & 65.2 & 37.28 & \textbf{47.44} & 5.56\\
BRENDA\_ribozyme &  0.0 & 0.0 & 0.0 & 0.55 & 0.0 & 0.0 & 0.0 & 0.51 & 8.24 & 1.11 & 1.96 & 0.48 & 18.24 & 2.46 & 4.34 & 0.86 & 26.39 & 25.68 & 26.03 & 2.24 & 30.75 & 31.16 & \textbf{30.95} & 2.67\\
OPE &  24.44 & 12.87 & 16.86 & 1.02  & 30.0 & 15.79 & 20.69 & 0.99 & 15.56 & 8.19 & 10.73 & 1.85 & 24.44 & 12.87 & 16.86 & 2.92 & 13.22 & 36.18 & \textbf{19.37} & 4.84 & 10.66 & 34.21 & 16.25 & 4.40\\ 
PPE & 10.0 & 3.7 & 5.41 & 1.75 &  0.0 & 0.0 & 0.0 & 1.21 & 3.33 & 1.23 & 1.80 & 2.11 & 23.33 & 8.64 & 12.61 & 3.55 & 39.39 & 64.20 & 48.83 & 11.61 & 52.5 & 77.78 & \textbf{62.69} & 14.09 \\
SE & 0.0 & 0.0 & 0.0 & 1.9 & 0.0 & 0.0 & 0.0 & 1.9 &  10.0 & 0.41 & 0.78 & 2.44 & 10.0 & 0.41 & 0.78 & 3.12 & 32.83 & 35.53 & 34.12 & 11.27 & 49.41 & 41.74 & \textbf{45.25} & 13.08\\
AE & 0.0 & 0.0 & 0.0 & 4.26 & 0.0 & 0.0 & 0.0 & 4.26 & 0.0 & 0.0 & 0.0 & 3.44 & 0.0 & 0.0 & 0.0 & 5.08 &  23.94 & 19.07 & \textbf{21.23} & 3.43 & 17.58 & 13.56 & 15.31 & 6.13\\
SuperMat & 50.0 & 2.45 & 4.67 & 2.12 &  0.0 & 0.0 & 0.0 & 1.91 & 7.14 & 0.35 & 0.67 & 2.71 & 10.71 & 0.52 & 1.0 & 3.63 & 23.40 & 7.69 & 11.58 & 9.27 &51.6 & 18.10 & \textbf{26.80} & 11.44 \\
\bottomrule
    \end{tabular}
    }
    \caption{Evaluation results across datasets comparing \textsc{SciLire} with other data generation tools. All scores are reported on a 0–100 scale, with the best F$_1$ score highlighted in \textbf{boldfaced}. \textsc{SciLire} results are based on GPT-5.}
    \label{tab:comparison_tools}
\end{table*}

\begin{table*}
\resizebox{\textwidth}{!}
{
\begin{tabular}{l| rrrr| rrrr| rrrr | rrrr }\toprule
& \multicolumn{8}{c|}{\textbf{SciSpace}} & \multicolumn{8}{c}{\textbf{ \textsc{SciLire} best (ICL large)}} \\\midrule
&  \multicolumn{4}{c|}{\textbf{Zero-shot}} &  \multicolumn{4}{c|}{\textbf{ICL (Static)}} &  \multicolumn{4}{c|}{\textbf{Zero-shot}} &  \multicolumn{4}{c}{\textbf{ICL (Dynamic)}} \\\midrule
        \bf Dataset & P & R & F$_1$ & ChrF & P & R & F$_1$ & ChrF & P & R & F$_1$ & ChrF & P & R & F$_1$ & ChrF \\\midrule
TDMS &  0.08 & 0.03 & 0.04  & 2.22 & 0.0 & 0.0 & 0.0 & 3.02 & 6.95 & 18.75 & 10.14 & 11.94 & 18.12 & 40.38 & \textbf{25.02} & 15.05\\
SciREX &   0.00  & 0.00  & 0.0  & 1.3 & 0.0 & 0.0 & 0.0 & 1.91 & 2.8 & 5.29 & 3.66 & 7.16 & 14.52 & 24.61 & \textbf{18.27} & 11.63\\
MPEA &  50.32  & 8.60  & 14.69  & 0.1 & 47.62 & 8.14 & 13.9 & 0.12 &  31.12 & 27.56 & 29.23 & 0.74 & 31.92 & 29.45 & \textbf{30.64} & 0.74\\
Diffusion &  18.60  & 0.49  & 0.95  & 0.27 & 14.67 & 0.38 & 0.75 & 0.3 & 33.51 & 11.86 & \textbf{17.52} & 1.54 & 30.35 & 12.0 & 17.20 & 1.52\\
YSHEAY &  0.00  & 0.00  & 0.00 & 3.73 & 0.0 & 0.0 & 0.0 & 5.0 & 2.94 & 29.51 & 5.34 & 12.32 & 4.47 & 35.36 & \textbf{7.93} & 13.28\\
CCRMG  & 0.00  & 0.00  & 0.00  & 3.33 & 0.0 & 0.0 & 0.0 & 3.21 &  0.96 & 18.7 & 1.82 & 11.64 & 1.53 & 25.20 & \textbf{2.89} & 12.48\\
Doping &  0.00  & 0.00  & 0.00  & 2.94 & 0.0 & 0.0 & 0.0 & 3.36 & 20.66 & 4.62 & 7.55 & 10.35 & 22.32 & 9.12 & \textbf{12.95} & 14.57\\
MMD & 0.00  & 0.00  & 0.00  & 0.29 & 0.0 & 0.0 & 0.0 & 0.25 &2.98 & 0.45 & 0.78 & 1.08 & 37.5 & 4.82 & \textbf{8.54} & 2.12 \\
MRL & 2.00  & 0.20  & 0.37  & 0.24 &  1.79 & 0.18 & 0.33 & 0.3 & 3.54 & 1.20 & 1.80 & 1.12 & 4.25 & 1.43 & \textbf{1.82} & 2.01\\
PNCExtract &  20.32  & 3.76  & 6.34  & 1.43 & 15.91 & 2.94 & 4.97 & 2.16 & 30.61 & 31.7 & 31.14 & 9.66 & 44.83 & 42.42 & \textbf{43.59} & 11.50\\
PolyIE & 0.00  & 0.00  & 0.00  & 5.12 & 0.0 & 0.0 & 0.0 & 4.95 & 11.43 & 19.04 & 14.29 & 15.67 & 15.88 & 21.69 & \textbf{18.34} & 15.75\\
BRENDA\_enzyme & 10.72  & 0.39  & 0.76  & 0.42 & 10.12 & 0.37 & 0.72 & 0.54 & 36.2 & 21.0 & 26.58 & 4.14 & 53.08 & 27.92 & \textbf{36.59} & 5.06\\
BRENDA\_ribozyme & 1.41  & 0.13  & 0.24  & 0.27 & 0.79 & 0.07 & 0.13 & 0.31 & 14.08 & 10.06 & 11.74 & 2.09 & 22.47& 15.69 & \textbf{18.48} & 2.51\\
OPE & 27.67  & 11.29  & 16.03  & 0.71 & 30.77 & 12.55 & 17.83 & 0.91 & 16.33 & 35.34 & \textbf{22.33} & 5.78 & 16.03 & 35.65 & 22.12 & 5.88\\
PPE &  2.14  & 0.88  & 1.25  & 1.13 & 0.0 & 0.0 & 0.0 & 1.12 & 41.65 & 68.05 & 51.67 & 11.62 & 53.91 & 80.57 & \textbf{64.60} & 13.81\\
SE & 0.00  & 0.00  & 0.00  & 1.71 & 0.0 & 0.0 & 0.0 & 1.71 & 40.50 &  33.38 & 36.6 & 11.42 & 51.83 & 42.62 & \textbf{46.78} & 12.8\\
AE & 0.00  & 0.00 & 0.00  & 3.48 & 0.0 & 0.0 & 0.0 & 3.46 & 17.25 &  13.53 & \textbf{15.17} & 5.12 & 13.56 & 10.03 & 11.53 & 9.34\\
SuperMat & 23.77  & 2.59 & 4.68  & 3.13 & 0.0 & 0.0 & 0.0 & 3.41 & 11.48 & 3.76 & 5.66 & 9.30 & 34.63 & 11.39 & \textbf{17.14} &11.06\\
\bottomrule
    \end{tabular}
    }
    \caption{Evaluation results across the full datasets comparing \textsc{SciLire} and SciSpace. F$_1$ reported with 0–100 scale; best score is \textbf{boldfaced}. \textsc{SciLire} results are based on GPT-5.}
    \label{tab:all_comparison}
\end{table*}

\section{User Data}
\label{sec:appendix_interaction}

\begin{figure*}[h]
    \centering
    \includegraphics[width=\textwidth]{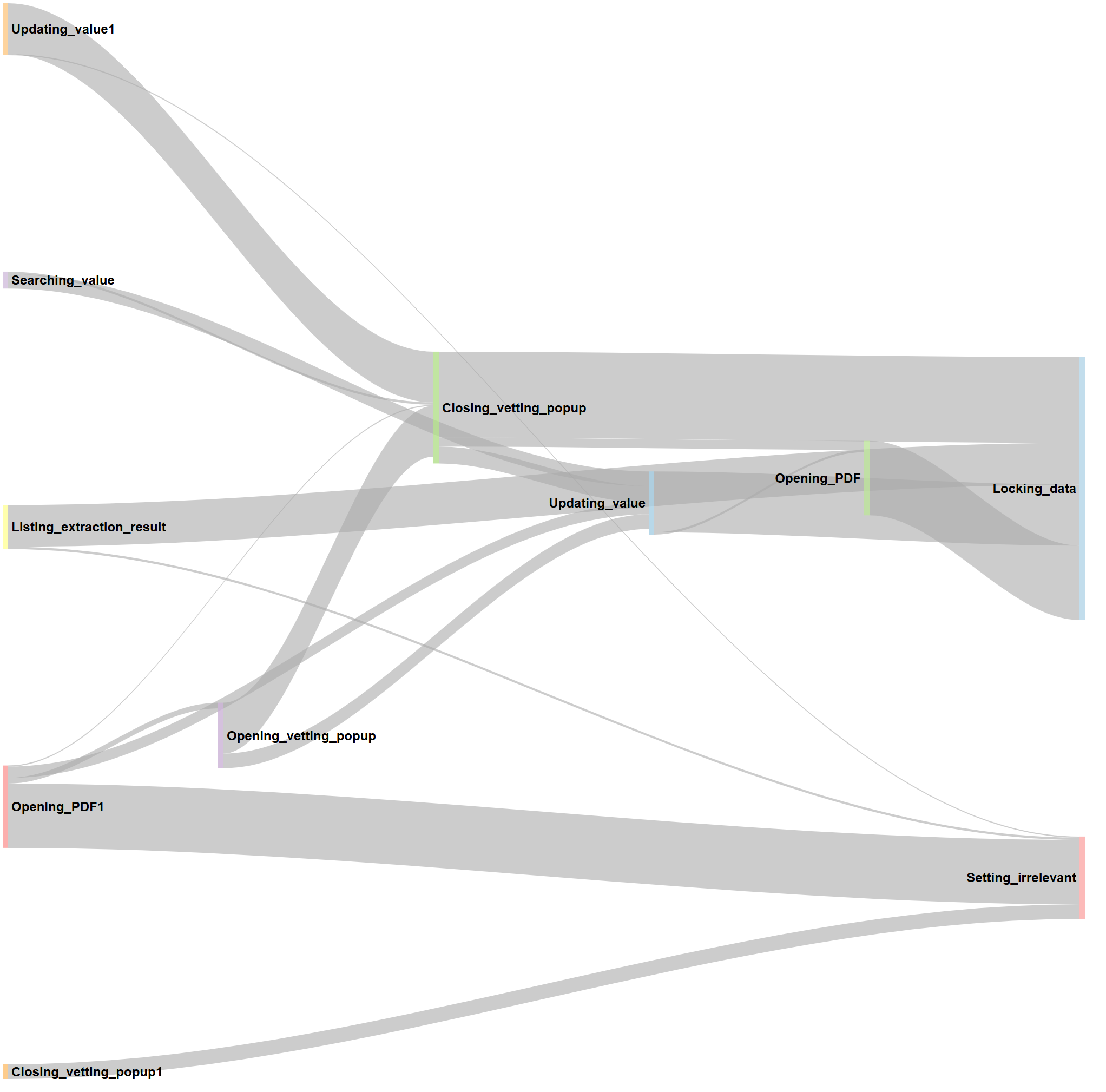}
    \caption{Interaction flows within \textsc{SciLire} for the early adopter trials.  Results show that most data acceptance (\textit{locking\_data}) or rejections (\textit{setting\_irrelevant}) occur via a data verification step (either checking the provenance data or the original source PDF). ``Vetting popup'' here refers to the verification support tools.  ``Updating\_value'' refers to human editing and manual data curation activities.  Actions with ``1'' at the end are used to eliminate cycles for the purposes of visualisation with a Sankey diagram.}
    \label{fig:interaction_flows}
\end{figure*}

In Figure~\ref{fig:interaction_flows}, we see that most of the acceptances and rejections of AI extracted records are performed via a verification step of either checking the tool's built-in verification support features or checking the reference.  This supports the premise that scientists wish to verify and curate data, consistent with our HAT-DC approach.  That pure automation is not necessarily what scientists are looking for.

\section{Demo Walkthrough}
\label{sec:appendix_demo_walkthrough}

In the demonstration software accompanying this paper, we consider the scenario where a user aims to curate a dataset for a well-known scenario in machine learning and computer science: generating a leaderboard in the computer science domain  (for an overview, see \citet{timmer-etal-2025-position}).

\begin{figure}
    \centering
    \frame{\includegraphics[width=\columnwidth]{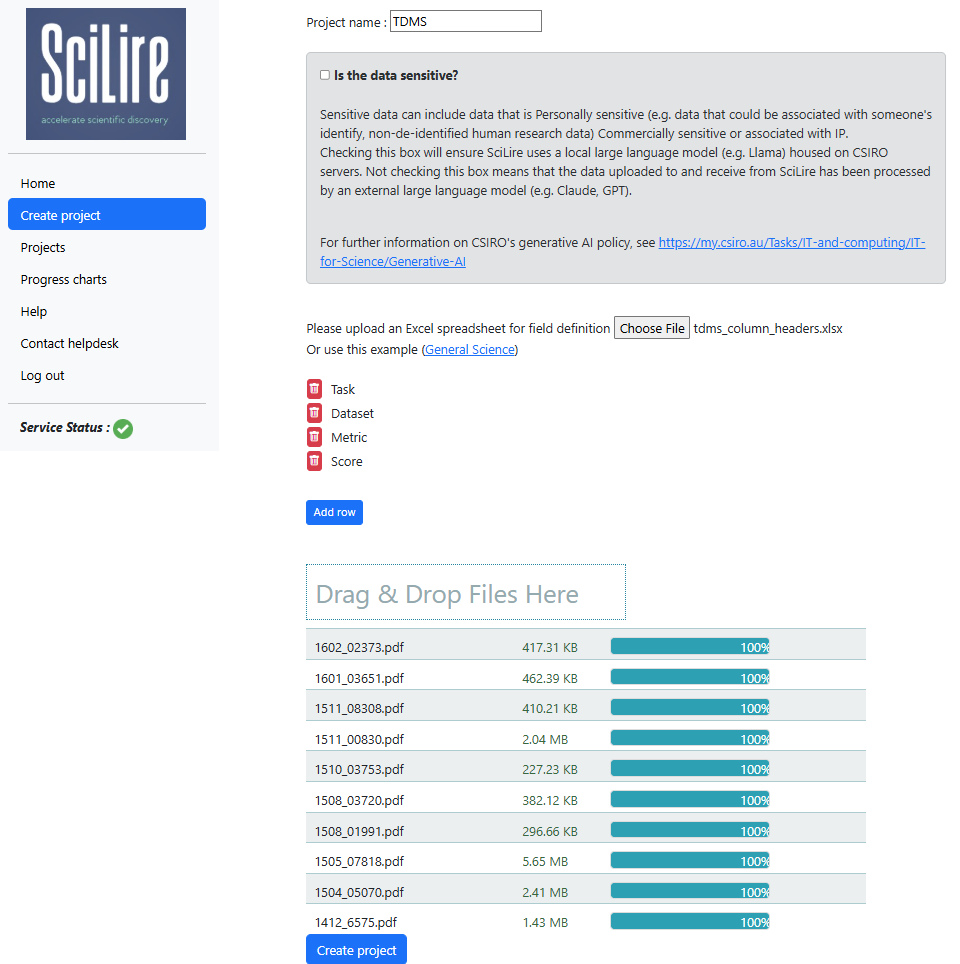}}
    \caption{A screenshot of \textsc{SciLire} for project creation.}
    \label{fig:scilire_project_create}
\end{figure}

Notionally, the user would first create a project by uploading the schema and a collection of documents (Figure~\ref{fig:scilire_project_create}). For demonstration purposes, papers have been uploaded and the project created in advance.\footnote{Due to legal constraints, we are unable to provide a non-licensed user account (i.e., a demo account) that demonstrates the uploading of the given dataset due to copyright legislation in Australia.}

\begin{figure*}[h]
    \centering
     \frame{\includegraphics[width=\textwidth]{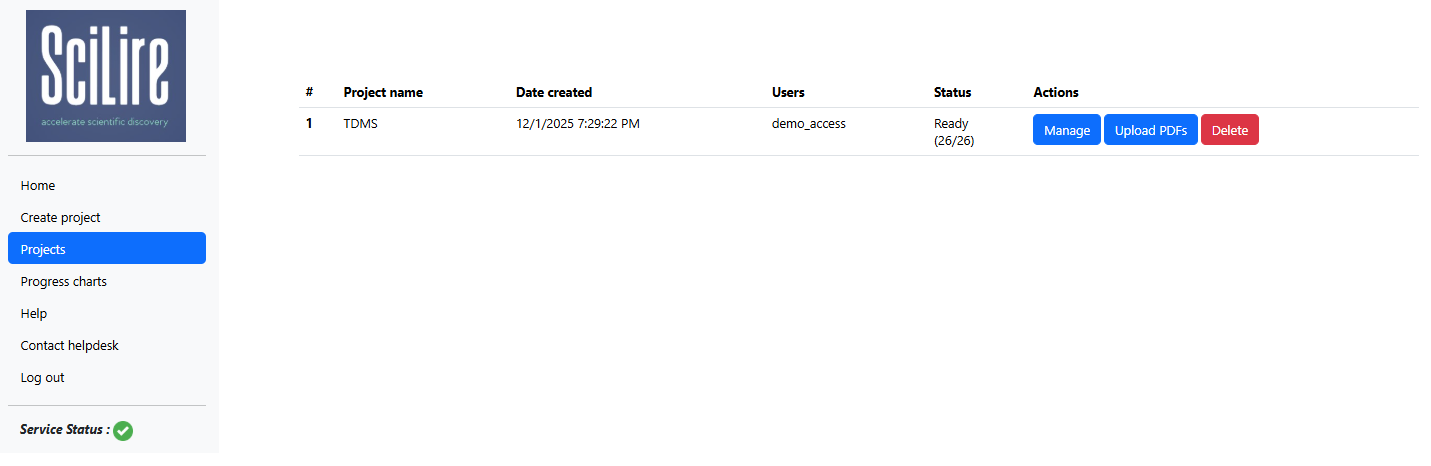}}
    \caption{A listing of a user's projects.}
    \label{fig:scilire_projects}
\end{figure*}

In Figure~\ref{fig:scilire_projects}, we see how the user navigates to their project.  By clicking on \textit{Projects} in the left-hand navigational menu, the user can click the option ``Manage'' for the registered project, here called ``TDMS'' (For the leaderboard dataset of the same name, which stands for ``Task, Dataset, Metric, Score''~\citep{hou2019identification}).  


\begin{figure*}[h]
    \centering
     \frame{\includegraphics[width=\textwidth]{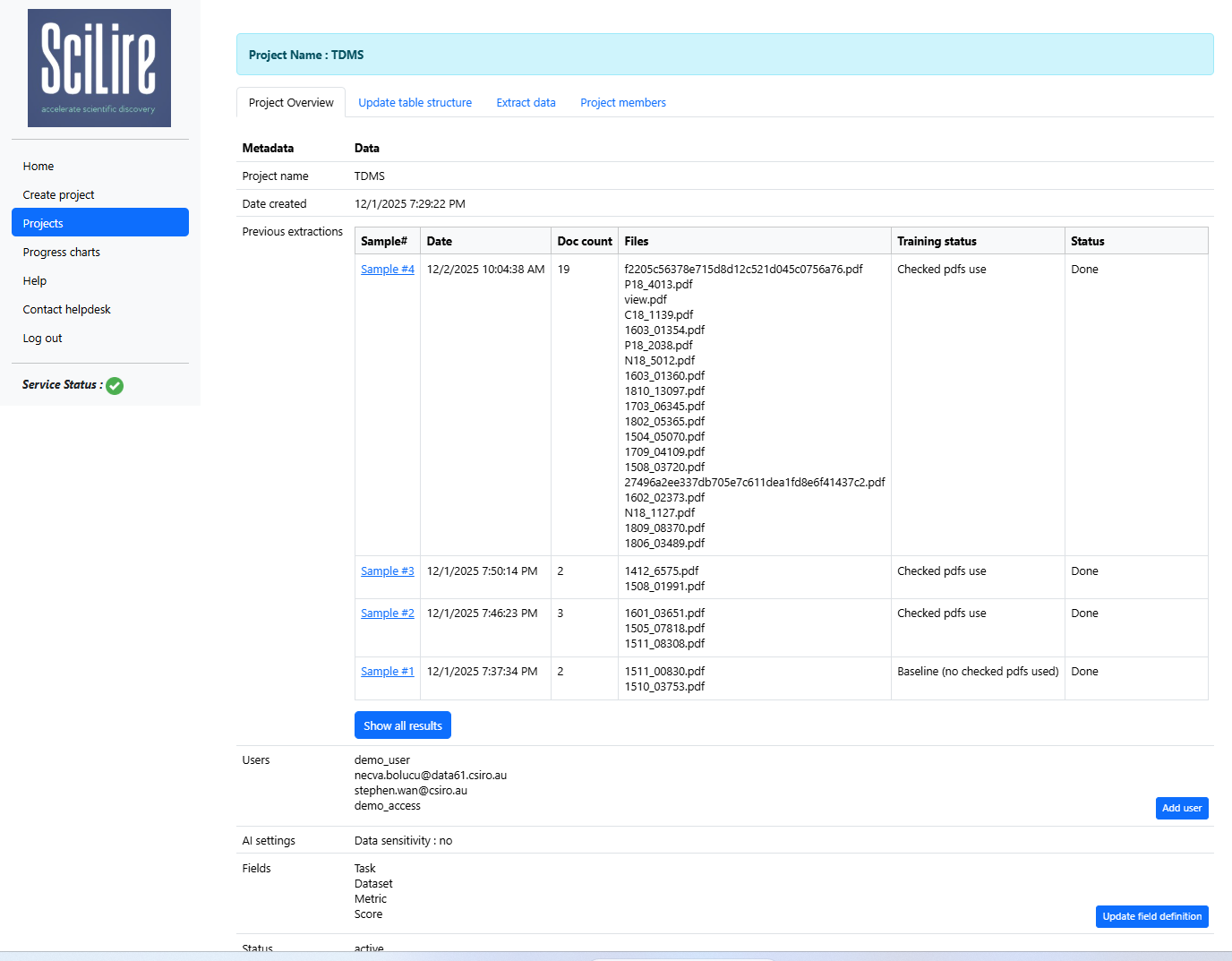}}
    \caption{The ``samples'' representing the table output from the iterative workflow.}
    \label{fig:scilire_samples}
\end{figure*}

In Figure~\ref{fig:scilire_samples}, we see how the user can revisit the outcomes of the Human-AI Team approach and the iterative data curation workflow.  The interface contains a table labelled ``Previous Extractions'', where Samples 1-3 represent iterations through the pilot phase.

\begin{figure*}[h]
    \centering
     \frame{\includegraphics[width=\textwidth]{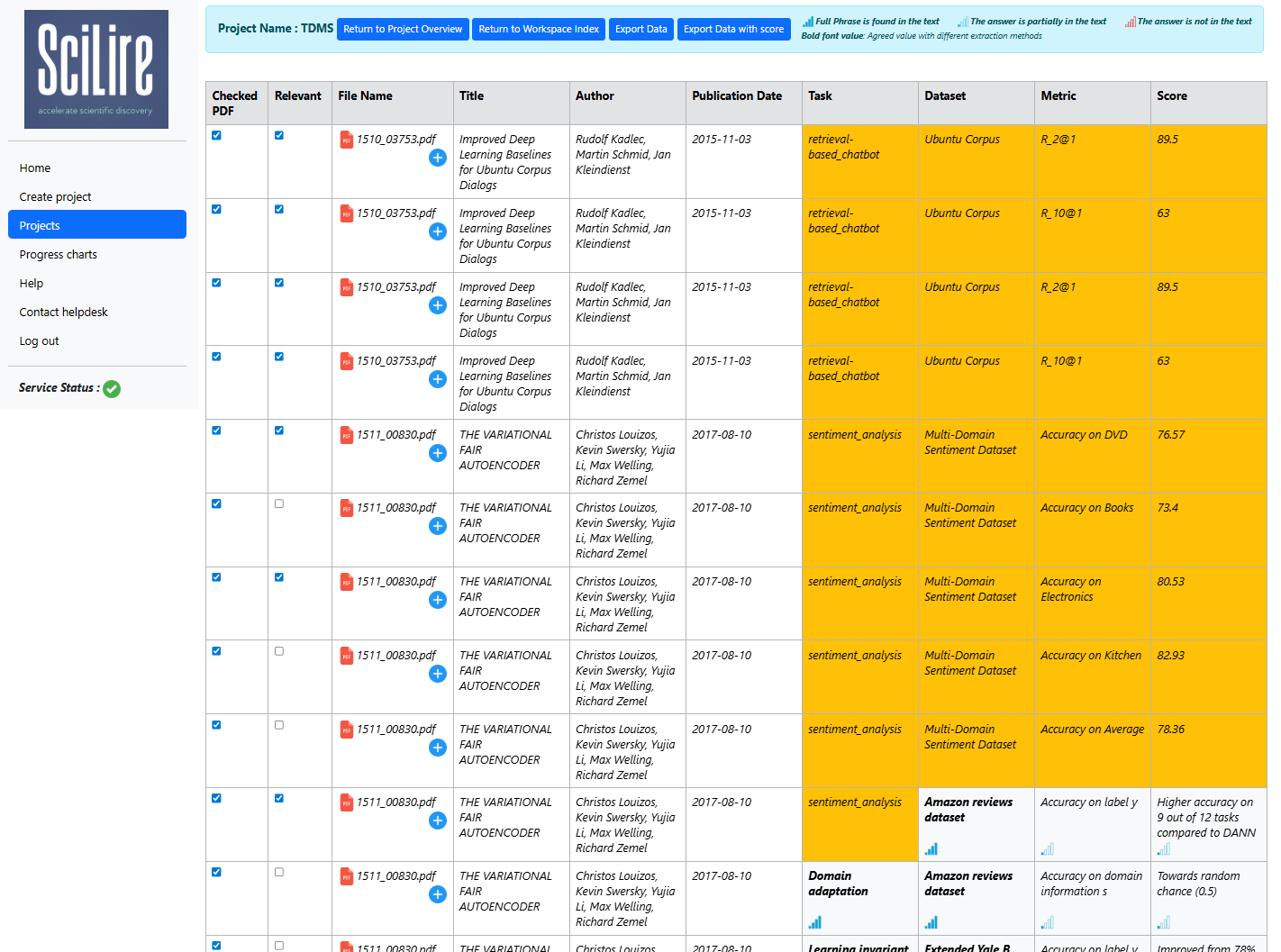}}
    \caption{The table in Sample 1, which has been edited and curated by the user.}
    \label{fig:scilire_sample1}
\end{figure*}

Figure~\ref{fig:scilire_sample1} shows a table in Sample 1, which has been edited and curated by the user.  Sample 1 is the first batch, in which \textsc{SciLire} generates records for the selected PDFs whereby the LLM generates results under the zero-shot setting. 
The yellow cells in Sample 1 show which data the user has reviewed and corrected. 

\begin{figure*}
    \centering
    \frame{\includegraphics[width=\textwidth]{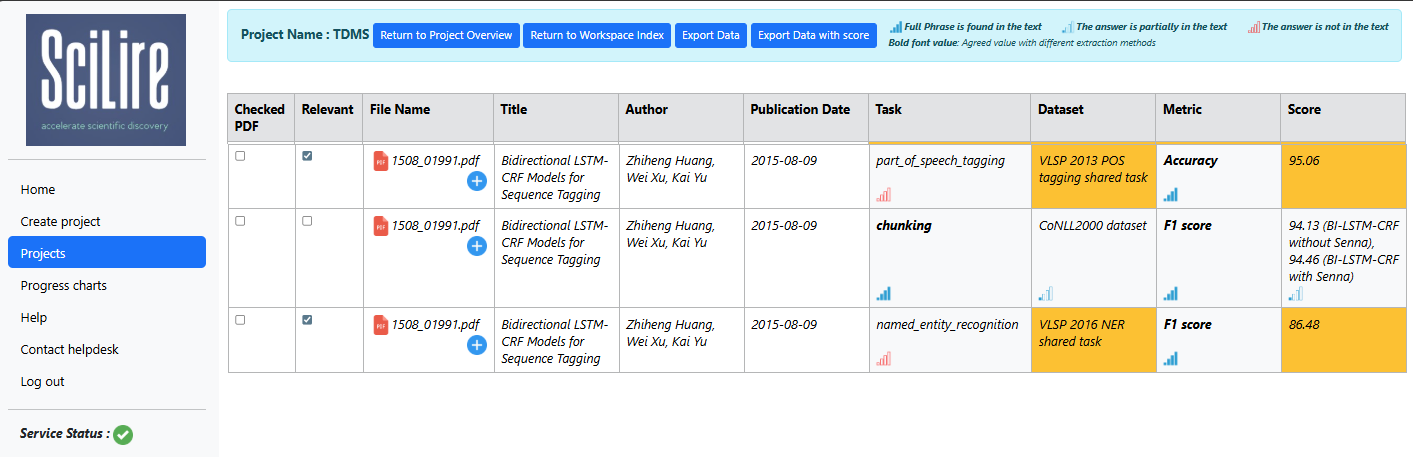}}
    \caption{A screenshot from the pilot phase (Sample 3) of the AI-augmented curation workflow in the demo.}
    \label{fig:first_phase}
\end{figure*}

Subsequent samples (e.g., Samples 2-3) proceed iteratively, with \textsc{SciLire} leveraging the user-corrected and verified records via dynamic sampling to generate records (Figure~\ref{fig:first_phase}).

Once the user is satisfied, they trigger the batch phase, where all remaining documents are processed to generate records, producing a complete, curated dataset ready.  This is represented by the output in Sample 4, which benefits from dynamic sampling drawn from the pool of data in Samples 1-3.

\end{document}